\documentclass[11pt]{article}

\usepackage[final]{acl}

\usepackage{times}
\usepackage{latexsym}

\usepackage[T1]{fontenc}

\usepackage[utf8]{inputenc}

\usepackage{microtype}

\usepackage{inconsolata}

\usepackage{graphicx}

\usepackage{amsmath}
\usepackage{amssymb}
\usepackage{amsthm}
\usepackage{mathtools}
\usepackage{bm}
\usepackage{booktabs}
\usepackage{multirow}
\usepackage{xcolor}
\usepackage{tcolorbox}
\usepackage{arydshln}
\usepackage{enumitem}

\title{Learning More from Less: Exploiting Counterfactuals for Data-Efficient Chart Understanding}

\author{
 \textbf{Jianzhu Bao\textsuperscript{1}},
 \textbf{Haozhen Zhang\textsuperscript{1}},
 \textbf{Kuicai Dong\textsuperscript{1}},
 \textbf{Bozhi Wu\textsuperscript{1}},
\\
 \textbf{Sarthak Ketanbhai Modi\textsuperscript{1}},
 \textbf{Zi Pong Lim\textsuperscript{2}},
 \textbf{Yon Shin Teo\textsuperscript{2}},
 \textbf{Wenya Wang\textsuperscript{1}}\thanks{\ \ Corresponding Author}
\\
\\
 \textsuperscript{1}Nanyang Technological University,
 \textsuperscript{2}Aumovio Singapore Pte. Ltd.,
\\
 \small{
   {jianzhubao@gmail.com, wangwy@ntu.edu.sg}
 }
}

\begin{document}
\maketitle
\begin{abstract}

Vision-Language Models (VLMs) have demonstrated remarkable progress in chart understanding, largely driven by supervised fine-tuning (SFT) on increasingly large synthetic datasets.
However, scaling SFT data alone is inefficient and overlooks a key property of charts: charts are programmatically generated visual artifacts, where small, code-controlled visual changes can induce drastic shifts in semantics and correct answers.
Learning this counterfactual sensitivity requires VLMs to discriminate fine-grained visual differences, yet standard SFT treats training instances independently and provides limited supervision to enforce this behavior.
To address this, we introduce ChartCF, a data-efficient training framework designed to enhance counterfactual sensitivity.
ChartCF consists of: (1) a counterfactual data synthesis pipeline via code modification, (2) a chart similarity-based data selection strategy that filters overly difficult samples for improved training efficiency, and (3) multimodal preference optimization across both textual and visual modalities.
Experiments on five benchmarks show that ChartCF achieves superior or comparable performance to strong chart-specific VLMs while using significantly less training data.\footnote{
Code is available at \url{https://github.com/jianzhubao/ChartCF}.}

\end{abstract}

\section{Introduction}

Chart understanding is a critical capability for VLMs, serving as a cornerstone for automated data analysis, document understanding, and scientific research \cite{masry2025bigcharts,he2024distill,xu2024chartmoe, dong2025docresearcher,dong2025mmdocrag,masry2025chartgemma}.
Given a chart image, VLMs must accurately extract important values, identify underlying trends, and perform complex reasoning to answer user questions.
To rigorously evaluate this capability, the research community has developed comprehensive benchmarks such as ChartQA \cite{DBLP:conf/acl/MasryLTJH22}, ChartX \cite{DBLP:journals/tip/XiaYYLZCSYZ25}, and CharXiv \cite{DBLP:conf/nips/WangXH0LZLWLMCA24}.
Despite the rapid progress of proprietary models like GPT-4o, open-source chart-specific VLMs still exhibit significant performance gaps on these challenging benchmarks.
They often struggle to locate subtle yet critical details and values in charts, leading to inferior extraction and reasoning performance compared to their proprietary counterparts.
Closing this gap is particularly important for real-world applications that involve sensitive data (e.g., financial reports, medical records) or large-scale chart processing, where locally deployable open-source VLMs are often preferred over proprietary APIs for privacy and cost considerations.

\begin{figure}
\centering
\includegraphics[width=0.45\textwidth]{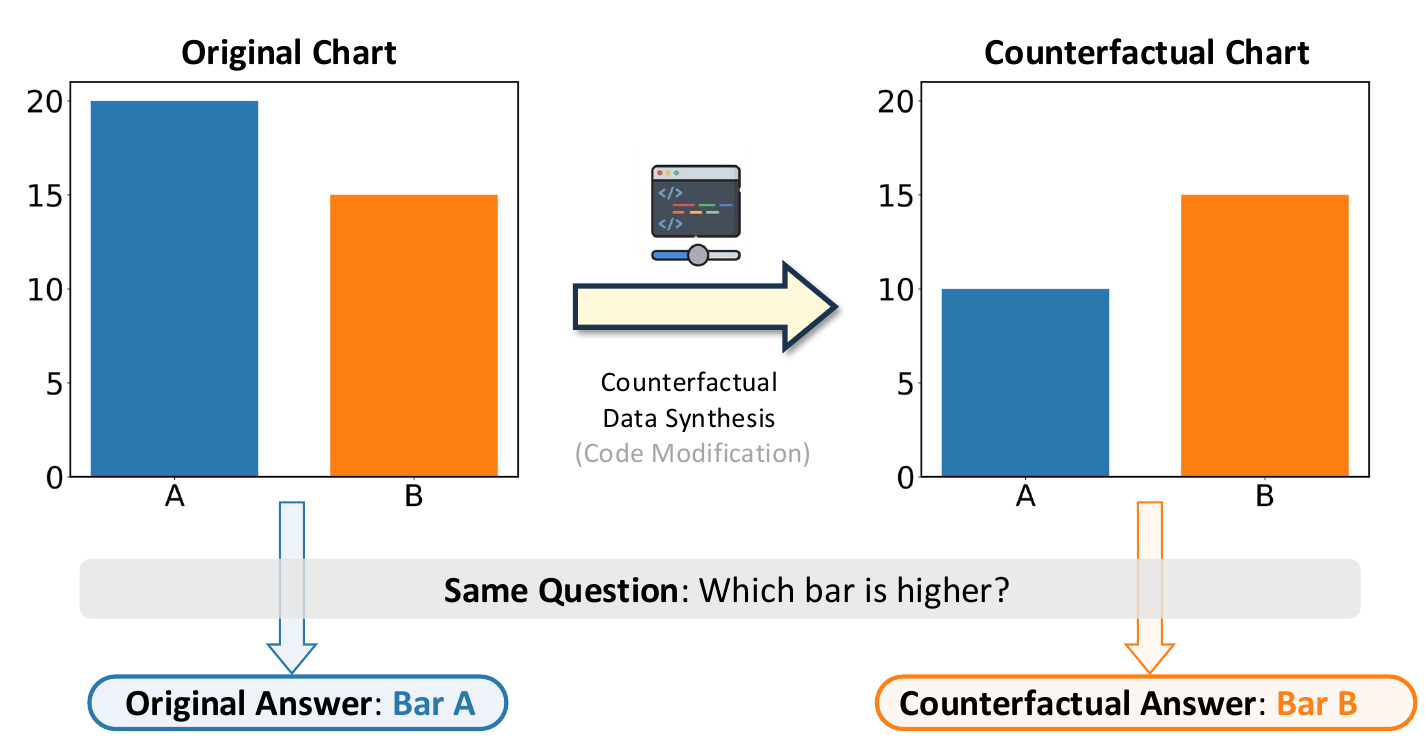}
\caption{\label{fig:1}Illustration of a counterfactual pair. Given the same question ``\textit{Which bar is higher?}'', a small modification to the height of bar A (left: original chart, right: counterfactual chart) results in a different answer.} 
\end{figure}

The prevailing approach to addressing this gap has focused on scaling up training data through sophisticated synthesis pipelines.
Recent work \cite{yang2025effective,he-etal-2025-distill} leverages code-based rendering tools (e.g., Matplotlib) and advanced LLMs to synthesize large-scale data for chart question answering, covering diverse topics and visual styles.
While supervised fine-tuning (SFT) on these datasets effectively improves VLMs' chart understanding capability, it is data-inefficient and fundamentally overlooks a unique property of charts that demands more targeted supervision.
Unlike natural images in other multimodal tasks \cite{DBLP:conf/emnlp/WangZHXZPC24,xiao2025exploring}, charts are programmatically generated visual artifacts where code controls all visual elements: from data values and trends to colors, labels, styles, etc. A small modification to any answer-critical element can alter the visual display and shift the semantic interpretation. This ultimately leads to a different correct answer.
As illustrated in Figure \ref{fig:1}, reducing the height of bar A causes the answer to ``Which bar is higher?'' to change from A to B,
despite the overall visual similarity between the two charts.
This \textit{counterfactual sensitivity} requires VLMs to discriminate fine-grained visual differences. However, standard SFT treats each chart-question pair independently, providing no explicit supervision for such discriminative behavior. Consequently, even VLMs trained on massive SFT data may hallucinate when subtle visual differences are critical.

To address this limitation, we introduce counterfactual supervision for chart understanding. Here, a counterfactual refers to a chart instance obtained through a minimal, code-controlled modification that preserves overall appearance while inducing a different correct answer. Building on this idea, we propose ChartCF, a data-efficient training framework designed to leverage such counterfactual chart pairs through preference optimization. It consists of three key components: (i) a counterfactual data synthesis pipeline, (ii) a chart similarity-based data selection strategy, and (iii) a contrastive preference optimization method.

First, to construct counterfactual data, we leverage existing high-quality synthetic datasets \cite{yang2025effective} and employ an advanced VLM to programmatically modify only answer-critical elements in the underlying plotting code.
This ensures that counterfactual charts remain visually similar while inducing different ground-truth answers.
Second, to improve training efficiency, we introduce a chart similarity-based data selection strategy that filters overly difficult samples, as they can introduce noise during preference learning.
Third, we apply Direct Preference Optimization~\cite{rafailov2023direct} across both modalities: Text DPO trains the model to favor answers corresponding to the presented chart while rejecting answers of counterfactual charts, whereas Image DPO associates each answer with its correct chart. This joint optimization effectively grounds answers in precise visual evidence.

We validate ChartCF on 5 widely-adopted benchmarks and results show that ChartCF achieves superior or comparable performance to strong open-source chart-specific VLMs.
Notably, compared to the ECD baseline~\cite{yang2025effective} trained on 300K samples, our approach achieves comparable performance with only 4K preference pairs.

\begin{itemize}[leftmargin=*, itemsep=-0.3em, topsep=0.0em]

    \item We introduce a data synthesis pipeline to generate \textit{counterfactual chart} pairs, coupled with a similarity-based selection strategy for improved training efficiency.
    
    \item We explore multimodal preference learning as a data-efficient alternative to SFT for chart understanding, demonstrating its effectiveness across multiple optimization objectives.

    \item Extensive experiments demonstrate that ChartCF achieves strong results while using significantly less training data, establishing a new paradigm for data-efficient chart understanding.
\end{itemize}

\section{Related Work}

Chart understanding \cite{huang2024pixels} encompasses several prominent tasks, such as chart question answering \cite{DBLP:conf/cvpr/KaflePCK18,DBLP:conf/wacv/MethaniGKK20,DBLP:conf/acl/MasryLTJH22,DBLP:conf/acl/MasryIABKKLRRST25}, chart-to-code generation \cite{DBLP:conf/acl/0001PKPLJACE23,DBLP:conf/acl/ZhaoL0000025,DBLP:conf/iclr/0002SLS0JXZLZLN25}, captioning \cite{DBLP:conf/inlg/ObeidH20,DBLP:conf/acl/KantharajLLMTHJ22}, and retrieval \cite{dong-etal-2025-mmdocir}.
Among these, chart question answering has emerged as a focal point of research due to its comprehensive nature, requiring models to integrate both visual perception and complex reasoning \cite{DBLP:conf/cvpr/KaflePCK18,wu2023dcqa,DBLP:conf/nips/WangXH0LZLWLMCA24,he-etal-2025-distill}. Consequently, we primarily focus on chart question answering in this paper.

In recent years, a surge of benchmarks has been introduced to evaluate the chart question answering task \cite{DBLP:conf/wacv/MethaniGKK20,DBLP:conf/emnlp/KantharajDLTHJ22,DBLP:journals/tip/XiaYYLZCSYZ25,xu2024chartbenchbenchmarkcomplexvisual}. These datasets generally fall into two categories: real-world and synthetic. Real-world benchmarks, such as ChartQA \cite{DBLP:conf/acl/MasryLTJH22} and CharXiv \cite{DBLP:conf/nips/WangXH0LZLWLMCA24}, provide authentic and diverse visualizations sourced from business reports and scientific publications, capturing the complexity of human-designed charts.
Conversely, synthetic benchmarks like PlotQA \cite{methani2020plotqa} and ReachQA \cite{he-etal-2025-distill} leverage programmatic tools and LLMs to generate diverse charts. 
These benchmarks have evolved from descriptive tasks to sophisticated multi-step reasoning over visual elements.

From a methodological perspective, a major research trend focuses on developing specialized chart models through large-scale training data synthesis. By leveraging code-based rendering tools and advanced LLMs, researchers have curated massive training datasets that emphasize both scale and diversity \cite{xu2024chartmoe,fan2025indepthinbreadthpretrainingmultimodal,masry2025bigchartsr1enhancedchartreasoning,tang2025visualprogrammabilityguidecodeasthought,DBLP:conf/aaai/HuangLZW0ZL25,huang2025evochart}. By applying SFT on these datasets, various chart-specific VLMs have been developed, exhibiting strong domain-specific performance \cite{han2023chartllama,DBLP:conf/emnlp/ZhangHXYXJZ024,DBLP:conf/acl/MasrySPHJ24,DBLP:conf/cikm/Jiang025,masry2025chartgemma,he-etal-2025-distill,yang2025effective}. 
Some recent work has explored reinforcement learning to further enhance reasoning capabilities \cite{huang2025chartsketcherreasoningmultimodalfeedback,sinha2025chartrvrreinforcementlearningverifiable}.
However, these approaches typically still require extensive SFT training on large-scale datasets before RL \cite{chen2025chartr1chainofthoughtsupervisionreinforcement,liu2025startspatialtextuallearning}.
In parallel, another line of research explores training-free approaches that leverage prompt engineering and external tools to enhance chart understanding \cite{wang2025chartagentchartunderstandingframework,kaur2025chartagentmultimodalagentvisually}.

Unlike these approaches that focus on scaling up training data, ChartCF emphasizes supervision quality through counterfactual data.
By providing explicit contrastive supervision on visually similar charts with different answers, combined with similarity-based data selection, ChartCF achieves effective training with significantly less data.

\section{Method}

\begin{figure*}
\centering
\includegraphics[width=0.9\textwidth]{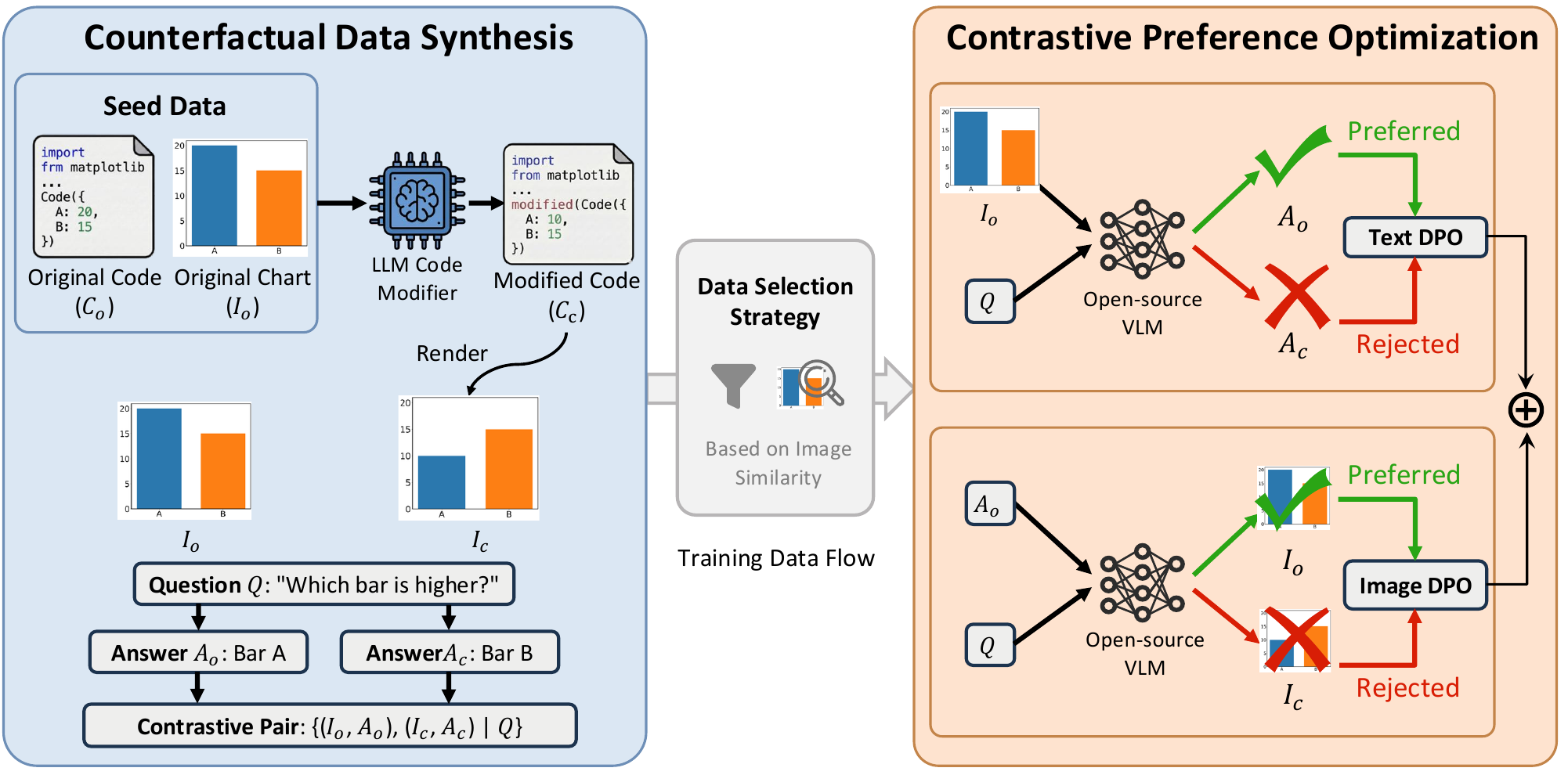}
\caption{\label{fig:2}Overview of ChartCF. \textbf{Left}: Counterfactual data synthesis pipeline that generates counterfactual chart pairs $(I_o, I_c)$ with their respective answers $A_o$ and $A_c$ via code-level modification. \textbf{Middle}: Chart similarity-based data selection that filters overly difficult pairs. \textbf{Right}: Contrastive preference optimization using Text DPO and Image DPO to train the model to distinguish between visually similar charts with different answers.}
\end{figure*}

We present ChartCF, a framework designed to boost the performance of VLMs in chart understanding through counterfactual-oriented training.
Figure \ref{fig:2} shows the architecture of ChartCF.
It first constructs a visual contrastive dataset through an automated counterfactual data synthesis pipeline, followed by a simple and efficient data selection strategy. 
Finally, ChartCF performs preference optimization targeting both textual responses and the chart images, improving the VLMs' capability to correctly associate visually similar charts with their respective answers.

\subsection{Problem Definition}

The goal of chart question answering is to generate a textual response $y$ given a chart image $I$ and a natural language question $x$. Formally, a VLM models the conditional probability $P_\theta(y | I, x)$, where $\theta$ represents the model parameters.

\subsection{Counterfactual Data Synthesis}
\label{subsec:counterfactual_data_synthesis}

To construct training data for preference optimization, we propose a counterfactual data synthesis pipeline. Instead of generating charts from scratch, we leverage the existing high-quality synthetic dataset, ECD \cite{yang2025effective}, to create visually similar contrastive pairs with discriminative yet small differences through code-level modification.

\paragraph{Initial Data.} We start with a dataset of seed samples, where each sample is a quadruplet $S_{o} = (I_{o}, C_{o}, Q, A_{o})$. Here, $I_{o}$ is the chart image, $C_{o}$ is the executable plotting code (e.g., Python Matplotlib script) used to render $I_{o}$, $Q$ is a question related to the chart, and $A_{o}$ is the corresponding ground-truth answer.

\paragraph{Counterfactual Code Generation.}
For each seed sample, we employ an advanced VLM (e.g., GPT-5) as a code modifier.
The goal is to generate a modified code snippet $C_{c}$ that results in a different answer for the same question $Q$, while maintaining overall visual similarity to the original chart.
Our prompt\footnote{Prompts are shown in Figures \ref{fig:descriptive_prompt} and \ref{fig:reasoning_prompt} of Appendix \ref{sec:prompt_details}.} design requires the advanced VLM to first isolate the specific code elements directly responsible for the current answer $A_{o}$—such as a specific data value, a subplot title, or a category label. 
The VLM is then instructed to modify only these answer-critical elements to produce $A_{c}$, while keeping all other components, including unrelated data points, colors, and random seeds, strictly identical to the original code $C_{o}$. 
This targeted modification process ensures that the difference between the original and counterfactual charts is isolated to the information required to answer the question, creating a precise signal for preference optimization.

\paragraph{Rendering and Pairing.} 
We execute the modified code $C_{c}$ to render the counterfactual chart image $I_{c}$, yielding a new sample $S_{c} = (I_{c}, C_{c}, Q, A_{c})$. 
Crucially, since the code modifications are strictly confined to answer-critical elements, $I_{c}$ maintains overall visual similarity to the original image $I_{o}$. 
By pairing the original sample $S_{o}$ with this counterfactual sample $S_{c}$, we construct a counterfactual pair:
\begin{equation}
    \mathcal{D}_{\text{pair}} = \{ (I_{o}, A_{o}), (I_{c}, A_{c}) | Q \}
\end{equation}
where $A_{c}$ is the ground-truth answer for $I_{c}$.
This synthesized data provides the necessary positive and negative signals for the subsequent preference optimization.
The quality of the synthesized data is further validated through a multi-stage pipeline (Appendix~\ref{sec:cost_and_yield}).

\subsection{Data Selection Strategy}
\label{subsec:data_selection_strategy}

Recent research suggests that preference learning algorithms, such as DPO, are sensitive to the difficulty level of training samples; overly difficult samples can prove detrimental rather than beneficial, potentially degrading optimization performance \cite{gou2025mixedpreferenceoptimizationreinforcement,DBLP:conf/icml/GaoL0XZ025}.
Inspired by this, we propose a simple yet effective data selection strategy based on chart image similarity.
Specifically, following previous work \cite{tang2025chartscodehierarchicalbenchmark,DBLP:conf/iclr/0002SLS0JXZLZLN25}, we employ GPT-5-mini to quantify the visual similarity score between the original chart $I_{o}$ and the counterfactual chart $I_{c}$.\footnote{The prompt is shown in Figure \ref{fig:similarity_prompt} of Appendix \ref{sec:prompt_details}.}
Intuitively, a higher similarity score implies that the visual modification is more subtle and thus harder to perceive, creating a hard negative sample.
Consequently, we rank the generated pairs by their image similarity scores and select the $\rho\%$ of samples with the lowest similarity scores (i.e., filtering out those samples with overly subtle visual changes).
This strategy eliminates ``noisy hard'' samples, allowing us to achieve superior data efficiency and model performance with a smaller, higher-quality dataset.

\subsection{Contrastive Preference Optimization}

We employ Direct Preference Optimization (DPO) \cite{rafailov2023direct} to align the model with the constructed counterfactual preference data.
Standard DPO typically focuses solely on optimizing preferences over textual responses.
However, we argue that for chart understanding, visual discrimination is equally critical.
Thus, following recent work on multimodal preference alignment \cite{DBLP:conf/emnlp/WangZHXZPC24,DBLP:conf/acl/WuSWH25}, we adopt a dual-alignment strategy that optimizes model preferences across both textual and visual modalities, formulated as a joint optimization of \textbf{Text DPO} and \textbf{Image DPO}. 

\paragraph{Text DPO.} 

Text DPO optimizes preferences at the textual response level.
Given the original chart image $I_{o}$ and question $Q$, the model should favor the ground-truth answer $A_{o}$ while rejecting the counterfactual answer $A_{c}$.
Note that $A_{c}$ is a particularly challenging negative sample because it is valid for the visually similar chart $I_{c}$ but factually incorrect for $I_{o}$.
We denote the policy model as $\pi_\theta$ and the reference model as $\pi_{\text{ref}}$. 
Following the standard DPO formulation \cite{rafailov2023direct}, the Text DPO loss is defined as:

{\small
\begin{multline}
    \mathcal{L}_{\text{text-dpo}}(I_{o}, Q, A_{o}, A_{c}) = \\
    - \log \sigma \left( \beta \log \frac{\pi_\theta(A_{o} | I_{o}, Q)}{\pi_{\text{ref}}(A_{o} | I_{o}, Q)} - \beta \log \frac{\pi_\theta(A_{c} | I_{o}, Q)}{\pi_{\text{ref}}(A_{c} | I_{o}, Q)} \right)
\end{multline}
}
where $\beta$ is the temperature parameter controlling the strength of the KL constraint.
By minimizing this loss, we penalize the model for generating answers that correspond to counterfactual visual states ($A_{c}$) when presented with the actual image ($I_{o}$), thereby sharpening its ability to distinguish fine-grained semantic differences.

\paragraph{Image DPO.}

Image DPO optimizes preferences at the visual input level, providing a complementary perspective to Text DPO.
It trains the VLM to identify the correct visual context for a given answer.
Specifically, we fix the question $Q$ and the original ground-truth answer $A_{o}$, and require the model to assign a higher likelihood to $A_{o}$ when conditioned on the correct image $I_{o}$ compared to the counterfactual image $I_{c}$.
The Image DPO loss is defined as:

{\small
\begin{multline}
    \mathcal{L}_{\text{img-dpo}}(I_{o}, I_{c}, Q, A_{o}) = \\
    - \log \sigma \left( \beta \log \frac{\pi_\theta(A_{o} | I_{o}, Q)}{\pi_{\text{ref}}(A_{o} | I_{o}, Q)} - \beta \log \frac{\pi_\theta(A_{o} | I_{c}, Q)}{\pi_{\text{ref}}(A_{o} | I_{c}, Q)} \right)
\end{multline}
}

\paragraph{Overall Training Objective.}

The final training objective combines the two preference losses:
\begin{equation}
    \mathcal{L}_{\text{total}} = \mathcal{L}_{\text{text-dpo}} + \mathcal{L}_{\text{img-dpo}}
\end{equation}
By jointly optimizing these two objectives, ChartCF trains the model to not only select the correct answer for a given chart but also associate each answer with its corresponding visual evidence.
We note that ChartCF's counterfactual data is broadly compatible with other multimodal preference optimization objectives, such as mDPO~\cite{DBLP:conf/emnlp/WangZHXZPC24} and S-VCO~\cite{DBLP:conf/acl/WuSWH25}, as shown in Section~\ref{sec:comparison_with_mdpo_svco}.

\section{Experiments}

\subsection{Experimental Setup}

\paragraph{Training Data.} 
We utilize the training data of ECD \cite{yang2025effective} as our data source, which provides over 10K chart images with Python plotting code and 300K QA pairs.
To construct our preference training set, we use all 10K chart images and randomly select one QA pair per image as seed samples.
We then employ gpt-5-2025-08-07 to modify the plotting code and generate a corresponding new answer.
This pipeline yields a training set of 10K preference pairs.
In the main experiments, we apply our data selection strategy to retain 4K pairs for training.
The impact of different data retention ratios is discussed in Section~\ref{sec:analysis_of_data_selection}.

\paragraph{Evaluation Benchmarks.}
We conduct experiments across five widely-adopted benchmarks, encompassing both real-world charts (CharXiv \cite{DBLP:conf/nips/WangXH0LZLWLMCA24}, ChartQA \cite{DBLP:conf/acl/MasryLTJH22}) and synthetic charts (ChartBench \cite{xu2024chartbenchbenchmarkcomplexvisual}, ChartX \cite{DBLP:journals/tip/XiaYYLZCSYZ25}, ECDBench \cite{yang2025effective}). Detailed descriptions of each benchmark are provided in Appendix~\ref{sec:benchmark_details}.
We primarily use CharXiv for detailed analysis experiments, due to its realism, complexity, and broad adoption.

\begin{table*}[t]
\small
\centering
\resizebox{0.88\textwidth}{!}{
\begin{tabular}{lcccccc}
\toprule
\multirow{2}{*}{Method} & \multirow{2}{*}{Model Size} & \multirow{2}{*}{\# Training Data} & \multicolumn{3}{c}{CharXiv} & \multirow{2}{*}{ChartQA} \\
\cmidrule{4-6}
 & & & Descriptive & Reasoning & Average & \\ \midrule
\multicolumn{6}{l}{\textcolor{gray}{\textit{Proprietary VLMs}}} \\
GPT-4o mini$^\dagger$ & - & - & 74.92 & 34.10 & 66.76 & 77.52 \\
GPT-4o$^\dagger$ & - & - & 84.45 & 47.10 & 76.98 & 85.70 \\
Claude-3.5-Sonnet$^\dagger$ & - & - & 84.30 & 60.20 & 79.48 & 90.80 \\
GPT-5 & - & - & 90.03 & 69.10 & 85.84 & 87.60 \\
Gemini-2.5-Pro & - & - & 91.22 & 68.40 & 86.66 & 89.68 \\
\midrule
\multicolumn{6}{l}{\textcolor{gray}{\textit{Chart-specific/Open-Source VLMs}}} \\
TinyChart$^\dagger$ & 3B & 1.36M (SFT) & - & 8.30 & - & 83.60 \\
ChartGemma$^\dagger$ & 3B & 123K (SFT) & 21.30 & 12.50 & 19.54 & 80.16 \\
ChartReasoner$^\dagger$ & 7B & 140K (SFT) & - & - & - & 86.93 \\
Chart-R1$^\dagger$ & 7B & 228K (SFT) + 30K (RL) & 62.00 & \underline{46.20} & 58.84 & \textbf{91.04} \\
ReachQA (InternVL2-8B)$^\dagger$ & 8B & 20K (SFT) & 54.83 & 32.70 & 50.40 & 82.44 \\
\hdashline
InternVL3.5 & 8B & - & 75.00 & 40.00 & 68.00 & 86.88 \\
\ \ \ \ + ECD & 8B & 300K (SFT) & 77.03 & 40.30 & 69.68 & 86.52 \\
\ \ \ \ + ChartCF (Ours) & 8B & \textbf{4K} pairs (DPO) & 80.47 & 42.30 & 72.84 & \underline{87.16} \\
Qwen2.5-VL$^\dagger$ & 7B & - & 66.40 & 41.20 & 61.36 & 83.04 \\
\ \ \ \ + ECD$^\dagger$ & 7B & 300K (SFT) & 74.20 & 40.20 & 67.40 & 85.32 \\
\ \ \ \ + ChartCF (Ours) & 7B & \textbf{4K} pairs (DPO) & 75.08 & 44.40 & 68.94 & 87.00 \\
Qwen3-VL & 8B & - & 81.25 & 43.10 & 73.62 & 85.44 \\
\ \ \ \ + ECD & 8B & 300K (SFT) & \underline{81.93} & 42.20 & \underline{73.98} & 85.12 \\
\ \ \ \ + ChartCF (Ours) & 8B & \textbf{4K} pairs (DPO) & \textbf{82.58} & \textbf{46.50} & \textbf{75.36} & 85.24 \\
\bottomrule
\end{tabular}
}
\caption{\label{tab:1}Performance comparison on real-world chart benchmarks: CharXiv and ChartQA. Results marked with $^\dagger$ are taken from prior work \cite{yang2025effective,he-etal-2025-distill,chen2025chartr1chainofthoughtsupervisionreinforcement}; other results are our own implementations. For chart-specific models, we report the number of training samples (\# Training Data), where SFT, RL, and DPO denote the amount of data used for SFT, reinforcement learning, and DPO, respectively. The best results among chart-specific and open-source VLMs are in \textbf{bold}, and the second-best are \underline{underlined}.}
\end{table*}

\paragraph{Baselines.}
We primarily compare against chart-specific VLMs, including ChartGemma \cite{masry2025chartgemma}, TinyChart \cite{DBLP:conf/emnlp/ZhangHXYXJZ024}, ChartReasoner \cite{jia2025chartreasonercodedrivenmodalitybridging}, Chart-R1 \cite{chen2025chartr1chainofthoughtsupervisionreinforcement}, ECD \cite{yang2025effective}, and ReachQA \cite{he-etal-2025-distill}.
Additionally, we evaluate general-purpose open-source VLMs as base models, including InternVL3.5-8B-Instruct \cite{wang2025internvl35advancingopensourcemultimodal}, Qwen2.5-VL-7B-Instruct \cite{Qwen2.5-VL}, Qwen3-VL-8B-Instruct \cite{qwen3technicalreport}, reporting their performance both before and after training with ChartCF.
Furthermore, we provide the performance of several proprietary models as reference points, including GPT-4o mini, GPT-4o \cite{openai2024gpt4ocard}, Claude-3.5-Sonnet \cite{claude}, GPT-5 \cite{gpt5}, and Gemini-2.5-Pro \cite{comanici2025gemini25pushingfrontier}.

\begin{table*}[t]
\centering
\resizebox{0.90\textwidth}{!}{
\begin{tabular}{lccccccccc}
\toprule
\multirow{2}{*}{Method} & \multirow{2}{*}{Model Size} & \multirow{2}{*}{\# Training Data} & \multicolumn{3}{c}{ChartBench} & \multirow{2}{*}{ChartX} & \multicolumn{3}{c}{ECDBench} \\
\cmidrule{4-6}\cmidrule{8-10}
 & & & Binary & NQA & Avg. &  & Des. & Rea. & Avg. \\ \midrule
\multicolumn{6}{l}{\textcolor{gray}{\textit{Proprietary VLMs}}} \\
GPT-4o mini$^\dagger$ & - & - & 70.26 & 34.93 & 74.33 & 44.36 & 57.27 & 24.26 & 40.77 \\
GPT-4o$^\dagger$ & - & - & 81.03 & 52.88 & 77.90 & 58.33 & 70.18 & 35.62 & 52.90 \\
Claude-3.5-Sonnet$^\dagger$ & - & - & 76.72 & 48.29 & 73.56 & 42.71 & 68.14 & 41.99 & 55.07 \\
GPT-5 & - & - & - & 79.33 & - & 83.51 & 78.27 & 63.15 & 70.71 \\
Gemini-2.5-Pro$^\dagger$ & - & - & - & 71.24 & - & 74.22 & 76.88 & 44.36 & 60.62 \\
\midrule
\multicolumn{6}{l}{\textcolor{gray}{\textit{Chart-specific/Open-Source VLMs}}} \\
ChartGemma$^\dagger$ & 3B & 123K (SFT) & 78.90 & 34.10 & 73.92 & 35.15 & - & - & - \\
ChartReasoner$^\dagger$ & 7B & 140K (SFT) & - & - & 55.20 & - & - & - & - \\
ReachQA (InternVL2-8B)$^\dagger$ & 8B & 20K (SFT) & 65.90 & 47.29 & 63.83 & 45.38 & - & - & - \\
\hdashline
InternVL3.5 & 8B & - & 79.42 & 67.71 & 78.12 & 68.49 & 49.67 & 32.27 & 40.97 \\
\ \ \ \ + ECD & 8B & 300K (SFT) & 79.00 & 67.33 & 77.70 & 70.31 & 65.77 & 39.79 & 52.78 \\
\ \ \ \ + ChartCF (Ours) & 8B & \textbf{4K} pairs (DPO) & 79.87 & 67.95 & 78.55 & 68.40 & 65.28 & 38.89 & 52.08 \\
Qwen2.5-VL$^\dagger$ & 7B & - & 80.99 & 67.81 & 79.53 & 67.80 & 57.35 & 19.04 & 38.19 \\
\ \ \ \ + ECD$^\dagger$ & 7B & 300K (SFT) & 79.35 & 70.86 & 78.41 & 70.83 & 66.34 & 35.38 & 50.86 \\
\ \ \ \ + ChartCF (Ours) & 7B & \textbf{4K} pairs (DPO) & \underline{82.45} & \textbf{72.43} & \underline{81.34} & \textbf{73.18} & 66.09 & 37.42 & 51.76 \\
Qwen3-VL & 8B & - & 71.43 & \underline{72.33} & 71.53 & 67.80 & \textbf{71.65} & 41.83 & \underline{56.74} \\
\ \ \ \ + ECD & 8B & 300K (SFT) & 81.42 & 71.95 & 80.37 & 71.61 & 70.51 & \textbf{43.22} & \textbf{56.87} \\
\ \ \ \ + ChartCF (Ours) & 8B & \textbf{4K} pairs (DPO) & \textbf{83.36} & 71.95 & \textbf{82.09} & \underline{72.14} & \underline{70.67} & \underline{42.08} & 56.38 \\
\bottomrule
\end{tabular}
}
\caption{\label{tab:2}Performance comparison on synthetic chart benchmarks: ChartBench, ChartX, and ECDBench. For ChartBench, ``Binary'' denotes yes/no questions and ``NQA'' denotes numerical questions. For ECDBench, ``Des.'' and ``Rea.'' refer to descriptive and reasoning questions, respectively. ``Avg.'' is short for ``Average''.}
\end{table*}

\paragraph{Evaluation Protocols.}

We follow ECD \cite{yang2025effective}\footnote{\url{https://github.com/yuweiyang-anu/ECD/tree/main/evaluation}} to ensure a fair comparison.
GPT-4o is employed to evaluate answer correctness against ground-truth references for all benchmarks except ChartBench ``yes/no'' questions, where we use regular expression matching.

\paragraph{Implementation Details.}

We employ Low-Rank Adaptation (LoRA) \cite{DBLP:conf/iclr/HuSWALWWC22} for efficient DPO fine-tuning.
Specifically, InternVL3.5-8B-Instruct, Qwen2.5-VL-7B-Instruct, and Qwen3-VL-8B-Instruct are fine-tuned with a LoRA rank of 64, alpha of 64, a batch size of 64, and a learning rate of 1e-4 for 2 epochs on 8 A100 80GB GPUs.
Following common practice in VLM fine-tuning, we freeze the vision encoder and projection layers, updating only the language model parameters during training.

\subsection{Main Results}

We present the main results on real-world and synthetic chart benchmarks in Tables~\ref{tab:1} and~\ref{tab:2}, respectively.
Overall, ChartCF consistently improves the performance of various base models (InternVL3.5, Qwen2.5-VL, and Qwen3-VL), achieving comparable or superior performance to strong open-source baselines across five benchmarks while using significantly less training data.
These results highlight that, compared to data-intensive SFT baselines, ChartCF achieves the most favorable trade-off between data efficiency and performance.

Specifically, among open-source models, ChartCF achieves the best average performance on CharXiv when applied to Qwen3-VL, particularly excelling on the most challenging reasoning questions (Table~\ref{tab:1}).
It also outperforms ECD on CharXiv across all three base models (Qwen2.5-VL, Qwen3-VL, InternVL3.5).
While ChartCF achieves lower scores than Chart-R1 on ChartQA, Chart-R1 requires extensive training with 228K SFT samples plus 30K RL samples, with training times of 3 hours for SFT and 30 hours for RL on 24 H800 GPUs \cite{chen2025chartr1chainofthoughtsupervisionreinforcement}.
In contrast, our method requires only 4K preference pairs for DPO training, which takes approximately 40 minutes on 8 A100 GPUs, resulting in substantially lower computational costs and training time.
According to Table \ref{tab:2}, on ChartBench, ChartX, and ECDBench, ChartCF achieves competitive or superior performance compared to ECD across various base models while using substantially less training data, further demonstrating data efficiency.

Beyond the standard ``Text DPO + Image DPO'' formulation, we further show in Section~\ref{sec:comparison_with_mdpo_svco} that ChartCF is compatible with alternative multimodal preference optimization objectives, with all variants achieving comparable performance.

\subsection{Enhancing SFT Baselines}

\begin{table}[t]
\small
\centering
\resizebox{\columnwidth}{!}{
\begin{tabular}{lcccc}
\toprule
\multirow{2}{*}{Method} & \multicolumn{3}{c}{CharXiv} & \multirow{2}{*}{ChartQA} \\
\cmidrule{2-4}
 & Des. & Rea. & Avg. \\ 
\midrule
\multicolumn{5}{l}{\textcolor{gray}{\textit{Qwen2.5-VL}}} \\
ChartCF Only (4K DPO) & 75.08 & 44.40 & 68.94 & \textbf{87.00} \\
ECD (300K SFT) & 74.20 & 40.20 & 67.40 & 85.32 \\
\ \ \ \ + ChartCF (300K SFT + 4K DPO) & \textbf{81.20} & \textbf{46.10} & \textbf{74.18} & 85.48 \\
\midrule
\multicolumn{5}{l}{\textcolor{gray}{\textit{Qwen3-VL}}} \\
ChartCF Only (4K DPO) & 82.58 & 46.50 & 75.36 & 85.24 \\
ECD (300K SFT) & 81.93 & 42.20 & 73.98 & 85.12 \\
\ \ \ \ + ChartCF (300K SFT + 4K DPO) & \textbf{83.45} & \textbf{48.40} & \textbf{76.44} & \textbf{86.76} \\
\bottomrule
\end{tabular}
}
\caption{\label{tab:enhancing_sft_baselines}Performance gains from applying ChartCF on top of ECD-trained models.}
\end{table}

To investigate whether ChartCF can further enhance VLMs already trained with large-scale SFT data, we apply our method on top of ECD-trained VLMs.
Results in Table~\ref{tab:enhancing_sft_baselines} show that ChartCF consistently improves performance across both Qwen2.5-VL and Qwen3-VL.
These results demonstrate that ChartCF provides complementary benefits to SFT, suggesting that counterfactual supervision and SFT capture different aspects of chart understanding.
This compatibility highlights ChartCF's practical value as a versatile enhancement for existing chart-specific VLMs.

\subsection{Ablation Studies}

\begin{table}[t]
\small
\centering
\resizebox{0.88\columnwidth}{!}{
\begin{tabular}{lcccc}
\toprule
\multirow{2}{*}{Method} & \multicolumn{3}{c}{CharXiv} & \multirow{2}{*}{ChartQA} \\
\cmidrule{2-4}
 & Des. & Rea. & Avg. \\ 
\midrule
ChartCF (Ours) & \textbf{75.08} & \textbf{44.40} & \textbf{68.94} & \textbf{87.00} \\
\ \ \ \ w/o Text DPO & 73.50 & 42.80 & 67.36 & 83.88 \\
\ \ \ \ w/o Image DPO & 72.67 & 43.00 & 66.74 & 84.56 \\
\ \ \ \ w/o Data Selection & 73.08 & 43.20 & 67.10 & 84.88 \\
\hdashline
SFT w. 4K Samples & 72.55 & 39.20 & 65.88 & 84.76 \\
SFT w. 8K Samples & 71.50 & 39.40 & 65.08 & 85.20 \\
\bottomrule
\end{tabular}
}
\caption{\label{tab:ablation}Ablation study on CharXiv and ChartQA based on Qwen2.5-VL. We evaluate the contribution of each component in ChartCF and compare against SFT baselines trained with similar amounts of data.
}
\end{table}

We conduct ablation studies on CharXiv and ChartQA using Qwen2.5-VL as the base model. Results are shown in Table~\ref{tab:ablation}.
First, removing Text DPO, Image DPO, or the data selection strategy individually leads to performance degradation across both benchmarks, demonstrating that dual optimization across both modalities combined with careful data selection is essential for ChartCF's success.
Second, we compare ChartCF against SFT baselines trained on equivalent data: ``SFT w. 4K Samples'' uses the original ECD samples corresponding to our 4K counterfactual pairs, while ``SFT w. 8K Samples'' additionally includes the 4K counterfactual samples, matching ChartCF's total training instances (4K pairs = 8K individual samples).
ChartCF consistently outperforms both SFT baselines, with particularly substantial improvements on reasoning questions, highlighting that explicit contrastive supervision is more effective than standard SFT.

\subsection{Analysis of Data Selection}
\label{sec:analysis_of_data_selection}

\begin{figure}[t]
\centering
\includegraphics[width=0.47\textwidth]{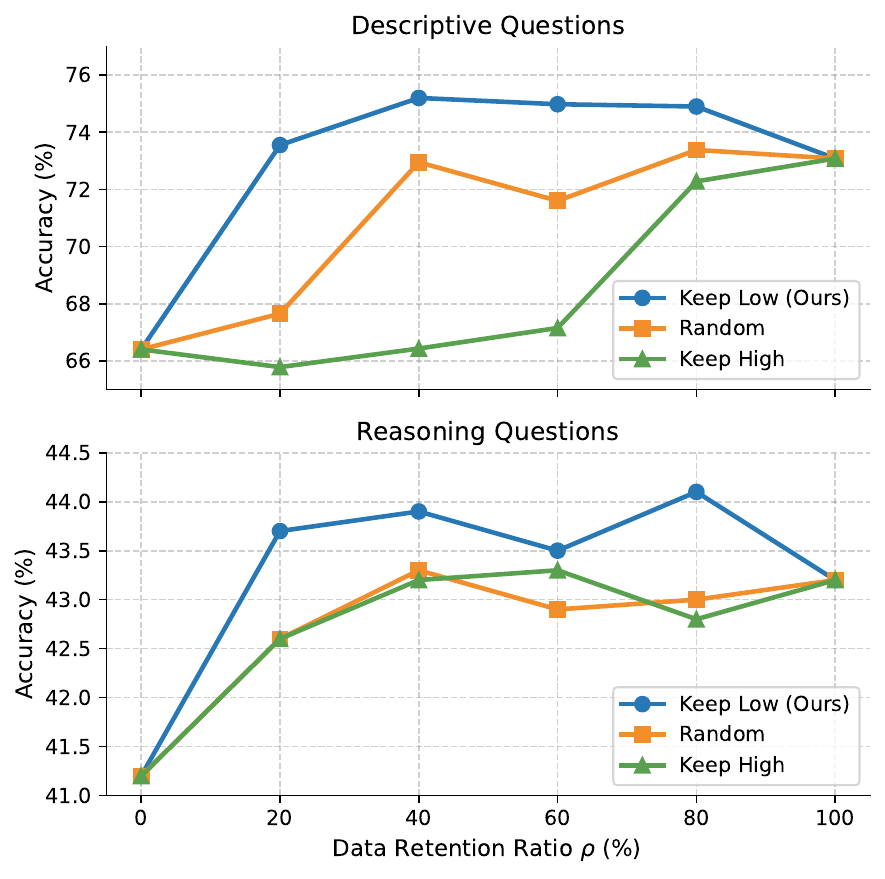}
\caption{\label{fig:3}Impact of data selection strategy on CharXiv. We compare three selection methods across different data retention ratios $\rho$ (x-axis):
``Keep Low (Ours)'' ranks counterfactual pairs by chart similarity and retains the $\rho\%$ pairs with the lowest similarity scores, ``Random'' randomly retains pairs, and ``Keep High'' retains pairs with the highest similarity scores.}
\end{figure}

To validate our chart similarity-based data selection strategy, we conduct experiments on CharXiv with varying data retention ratios $\rho$, as shown in Figure~\ref{fig:3}.
We compare our strategy (``Keep Low'') against two baselines: random sampling and an inverse strategy that retains pairs with the highest similarity scores (``Keep High'').

Our ``Keep Low'' strategy consistently outperforms random sampling and ``Keep High'' across various retention ratios, demonstrating that selecting pairs with lower similarity is essential for identifying valuable training samples.
Notably, our method achieves optimal performance when $\rho=40$, where only $\sim$4K pairs of training samples are included.
The performance of ``Keep Low'' first increases as more data is included, then decreases at higher retention ratios.
This reflects two competing effects: initially, too few samples lead to insufficient training, while at higher retention ratios, the inclusion of excessively difficult samples introduces noise that degrades performance.
In contrast, random sampling shows an increasing trend but fails to reach the peak performance of our method due to persistent noisy samples at all ratios.
Finally, ``Keep High'' performs poorly, especially on descriptive questions, as excessively difficult samples act as noise.
These results validate our hypothesis that preference learning is sensitive to sample difficulty.
Furthermore, filtering based on chart similarity provides an effective mechanism to balance training data quality and quantity.

\subsection{Analysis of Different Types of Hard Samples}
\label{subsec:different_types_of_hard_samples}

To better understand what makes a counterfactual pair difficult for preference optimization, we investigate two types of synthetic data.
The first type, ``Synthetic Data w/o Distractors'', follows our standard pipeline in Section \ref{subsec:counterfactual_data_synthesis}, modifying only answer-critical elements.
The second type, ``Synthetic Data w. Distractors'', additionally modifies non-critical visual elements (e.g., unrelated data points, colors) that do not affect the answer but increase overall visual differences.
For each type, we examine different chart similarity-based selections: the lowest half and the highest half.

Results in Table~\ref{tab:hard_sample_type} reveal two categories of hard samples that hinder DPO training.
First, within the clean synthetic data without distractors, high-similarity pairs (Highest Half) prove detrimental, significantly degrading performance compared to low-similarity pairs (Lowest Half).
This confirms our hypothesis in Section \ref{subsec:data_selection_strategy} that overly subtle modifications create noisy signals for training.
Second, adding distractors uniformly hurts performance across all similarity levels.
This suggests that distractors introduce spurious correlations that mislead the preference optimization process, as the model struggles to identify which visual differences are answer-critical versus merely distracting.
These findings highlight that effective preference learning requires counterfactual pairs that isolate answer-critical differences while remaining visually distinguishable, as both overly subtle changes and irrelevant modifications impede learning.

\begin{table}[t]
\small
\centering
\resizebox{0.88\columnwidth}{!}{
\begin{tabular}{lccc}
\toprule
\multirow{2}{*}{Method} & \multicolumn{3}{c}{CharXiv}\\
\cmidrule{2-4}
 & Des. & Rea. & Avg. \\ 
\midrule
Qwen2.5-VL & 66.40 & 41.20 & 61.36 \\
\hdashline
\multicolumn{4}{l}{\textcolor{gray}{\textit{Synthetic Data w/o Distractors}}} \\
ChartCF w/o Data Selection (10K) & 73.08 & 43.20 & 67.10 \\
\ \ \ \ w. Lowest Half Pairs (5K) & \textbf{75.65} & \textbf{43.70} & \textbf{69.26} \\
\ \ \ \ w. Highest Half Pairs (5K) & 66.42 & 43.30 & 61.80 \\
\hdashline
\multicolumn{4}{l}{\textcolor{gray}{\textit{Synthetic Data w. Distractors}}} \\
ChartCF w/o Data Selection (10K) & 69.92 & 43.10 & 64.56 \\
\ \ \ \ w. Lowest Half Pairs (5K) & 64.60 & 38.90 & 56.26 \\
\ \ \ \ w. Highest Half Pairs (5K) & 66.80 & 40.00 & 61.44 \\
\bottomrule
\end{tabular}
}
\caption{\label{tab:hard_sample_type} Analysis of different types of hard samples on CharXiv. We compare clean synthetic data (modifications only to answer-critical elements) versus data with distractors (additional modifications to non-critical elements), across different similarity-based selections.
}
\end{table}

\subsection{Compatibility with Different Code Modifiers}
\label{sec:different_code_modifiers}

To verify the compatibility of ChartCF with different code modifiers, we conduct a small-scale experiment using 2K counterfactual pairs synthesized by various code modifiers, including four proprietary VLMs (GPT-5, GPT-4o, Gemini-2.5-Pro, and Claude-4.5-Sonnet) and three open-source alternatives from the Qwen3-VL-Thinking family (8B, 32B, and 235B-A22B).
As shown in Table~\ref{tab:data_synthesis_with_different_vlms}, ChartCF yields consistently strong performance across all modifiers. Notably, even compact open-source models such as Qwen3-VL-8B-Thinking and Qwen3-VL-32B-Thinking achieve substantial gains over the base Qwen2.5-VL, and the larger Qwen3-VL-235B-A22B-Thinking closes the gap further, reaching performance competitive with proprietary alternatives such as GPT-5. These findings suggest that the effectiveness of ChartCF is primarily driven by the counterfactual data construction paradigm itself, rather than the capability of any specific proprietary model. Practitioners can adopt our pipeline with locally deployed open-source models, maintaining data privacy while achieving comparable performance at significantly lower cost.

\begin{table}[t]
\small
\centering
\resizebox{\columnwidth}{!}{
\begin{tabular}{lccc}
\toprule
\multirow{2}{*}{Method} & \multicolumn{3}{c}{CharXiv}\\
\cmidrule{2-4}
 & Des. & Rea. & Avg. \\ 
\midrule
Qwen2.5-VL & 66.40 & 41.20 & 61.36 \\
\hdashline
\multicolumn{4}{l}{\textcolor{gray}{\textit{Proprietary Code Modifiers}}} \\
\ \ \ \ + ChartCF w. GPT-5 & 73.55 & 43.70 & 67.58 \\
\ \ \ \ + ChartCF w. GPT-4o & 73.43 & 42.20 & 67.18 \\
\ \ \ \ + ChartCF w. Gemini-2.5-Pro & 75.00 & 43.30 & 68.66 \\
\ \ \ \ + ChartCF w. Claude-4.5-Sonnet & \textbf{75.88} & \textbf{44.00} & \textbf{69.50} \\
\hdashline
\multicolumn{4}{l}{\textcolor{gray}{\textit{Open-Source Code Modifiers}}} \\
\ \ \ \ + ChartCF w. Qwen3-VL-235B-A22B-Thinking & 74.35 & 42.90 & 68.06 \\
\ \ \ \ + ChartCF w. Qwen3-VL-32B-Thinking & 74.30 & 42.20 & 67.88 \\
\ \ \ \ + ChartCF w. Qwen3-VL-8B-Thinking & 72.35 & 41.80 & 66.24 \\
\bottomrule
\end{tabular}
}
\caption{\label{tab:data_synthesis_with_different_vlms}Performance comparison on CharXiv using different proprietary and open-source VLMs as code modifiers, trained on a 2K sample subset.}
\end{table}

\subsection{Compatibility with Alternative Multimodal Preference Optimization Objectives}
\label{sec:comparison_with_mdpo_svco}

Beyond the standard ``Text DPO + Image DPO'' formulation, ChartCF is broadly compatible with other multimodal preference optimization objectives designed for fine-grained visual grounding. To demonstrate this, we apply two representative methods, mDPO~\cite{DBLP:conf/emnlp/WangZHXZPC24} and S-VCO~\cite{DBLP:conf/acl/WuSWH25}, to our counterfactual data. Specifically, mDPO augments the ``Text DPO + Image DPO'' framework with an Anchored Preference Optimization (AncPO) loss, while S-VCO further introduces a symmetric visual contrastive term. For fairness, all methods are trained on Qwen2.5-VL using the same 4K counterfactual pairs as in our main experiments.

\begin{table}[t]
\small
\centering
\resizebox{\columnwidth}{!}{
\begin{tabular}{lcccc}
\toprule
\multirow{2}{*}{Method} & \multicolumn{3}{c}{CharXiv} & \multirow{2}{*}{ChartQA} \\
\cmidrule{2-4}
 & Des. & Rea. & Avg. & \\
\midrule
ECD (300K SFT) & 74.20 & 40.20 & 67.40 & 85.32 \\
\hdashline
ChartCF (Ours) & 75.08 & \textbf{44.40} & 68.94 & 87.00 \\
mDPO (ChartCF + AncPO) & 74.38 & 43.00 & 68.10 & \textbf{87.28} \\
S-VCO & \textbf{75.85} & 43.30 & \textbf{69.34} & 86.44 \\
VCO (S-VCO w/o sym.) & 75.25 & 41.10 & 68.42 & 86.04 \\
ChartCF + S-VCO & 74.55 & 44.00 & 68.44 & 87.08 \\
\bottomrule
\end{tabular}
}
\caption{\label{tab:mdpo_svco_comparison}Compatibility of ChartCF's counterfactual data with alternative multimodal preference optimization objectives on CharXiv and ChartQA, all trained on Qwen2.5-VL with the same 4K counterfactual pairs. ``VCO'' denotes S-VCO without the symmetric term. ``ChartCF + S-VCO'' combines the two losses.}
\end{table}

As shown in Table~\ref{tab:mdpo_svco_comparison}, ChartCF's counterfactual data works effectively with various preference optimization objectives, with all variants achieving comparable performance and consistently outperforming the 300K-sample SFT baseline.

\subsection{Case Study}

We present two types of case studies in the appendices.
First, we showcase four counterfactual chart pairs generated by our synthesis pipeline, covering descriptive questions (legend interpretation, axis scale recognition) and reasoning questions (numerical computation, spatial relationships), with modifications including textual labels, numerical values, and visual element positions (Appendix~\ref{sec:case_study_of_counterfactual_data_synthesis}).
Second, we compare ChartCF and ECD on five examples, demonstrating that ChartCF achieves superior performance on tasks requiring fine-grained element localization, precise value reading, spatial reasoning, and numerical computation (Appendix~\ref{sec:case_study_of_model_predictions}).

\subsection{Additional Analyses}

In the appendix, we provide additional analyses on the cost and validation of our data synthesis pipeline (Appendix~\ref{sec:cost_and_yield}), as well as a study on curriculum learning for hard samples (Appendix~\ref{sec:curriculum}).

\section{Conclusion}

We introduce ChartCF, a data-efficient training framework that enhances chart understanding through counterfactual learning.
Unlike conventional approaches that rely on scaling training data, our method leverages the programmatic nature of charts to synthesize targeted counterfactual pairs via code modifications.
Combined with similarity-based data selection and contrastive preference optimization across both textual and visual modalities, ChartCF achieves competitive or superior performance compared to strong chart-specific VLMs while using significantly less training data.

\section*{Limitations}

While ChartCF demonstrates strong data efficiency, our work has several limitations that warrant future investigation.

First, our approach relies on the availability of executable plotting code.
While synthetic chart datasets with code are increasingly available, extending our method to naturally occurring charts would require developing code generation or reverse-engineering techniques.

Second, our counterfactual synthesis pipeline depends on advanced VLMs (e.g., GPT-5) to generate modified code, which introduces additional costs during data preparation.
However, we show in Section~\ref{sec:different_code_modifiers} that open-source models can serve as effective alternatives, significantly reducing the overhead of our method.

\section*{Ethics Statement}

This work uses publicly available datasets that have been widely adopted in prior chart understanding research. Our use of these datasets, and all other software and resources, strictly complies with their respective licenses and intended purposes. The datasets are understood to be free of personally identifiable information and offensive content.
AI assistants are employed solely for grammar checking and text polishing during manuscript preparation.

We acknowledge that using LLMs for counterfactual data synthesis may introduce potential risks such as inaccuracies in generated code and answers. However, our experimental results demonstrate consistent performance improvements across multiple benchmarks, validating the effectiveness of our method despite potential noise in synthesized data. Moreover, synthetic data generation remains significantly more cost-effective and scalable than manual annotation.

\section*{Acknowledgments}

This research was performed at the AUMOVIO-NTU Corporate Lab.
This research/project is supported by A*STAR under the RIE2025 Industry Alignment Fund – Industry Collaboration Projects (IAF-ICP) Funding Initiative (Award: I2501E0045), as well as cash and in-kind contribution from the industry partner(s).
This research is supported in part by computing resources provided through the AWS Build on Trainium program.

\bibliography{custom}

\appendix

\clearpage
\section*{Appendix}

\section{Prompt Details}
\label{sec:prompt_details}

This section provides the detailed prompts used in this work.

\subsection{Prompts for Counterfactual Data Synthesis}

Due to the distinct nature of question types in existing datasets \cite{yang2025effective,he-etal-2025-distill}, we design two separate prompts for code modification: one for descriptive questions and another for reasoning questions. This differentiation is motivated by the fact that reasoning questions, unlike descriptive ones, require the model to generate not only the final answer but also the intermediate reasoning steps.

\begin{itemize}
    \item Figure~\ref{fig:descriptive_prompt}: Prompt for descriptive questions.
    \item Figure~\ref{fig:reasoning_prompt}: Prompt for reasoning questions, which includes instructions for generating intermediate reasoning steps.
\end{itemize}

\subsection{Prompt for Image Similarity Evaluation}

To implement our similarity-based data selection strategy (Section~\ref{subsec:data_selection_strategy}), we employ GPT-5-mini to evaluate the visual similarity between original and counterfactual chart pairs.

\begin{itemize}
    \item Figure~\ref{fig:similarity_prompt}: Prompt for evaluating image similarity scores.
\end{itemize}

\begin{figure*}[t]
\begin{tcolorbox}[
    colback=gray!10,
    colframe=gray!40,
    verbatim,
    fontupper=\footnotesize,
    ]
\scriptsize
**Task:** Given a chart image, its plotting code, a descriptive question, and the current answer, modify the code so that the answer to the question becomes different. You should ONLY modify the element(s) directly responsible for the current answer. \\
\\
\#\# Requirements \\
\\
- First assess whether you think you are capable of reasonably accomplishing this task \\
- Identify the specific data point(s) or element(s) that determine the current answer \\
- Modify ONLY those necessary elements to produce a different answer \\
- Do NOT change any other data points, labels, colors, or visual elements \\
- Do NOT change the final output/save path in the original code: it must remain `rendered\_images/\{6-digit-number\}.png', e.g., `rendered\_images/000002.png'. \\
- Do NOT modify the \texttt{\`{}}set\_random\_seed\texttt{\`{}} function or the random seed value it sets \\
- Ensure the modification is visually noticeable to human eyes (e.g., at least 15-25\% change for numerical values) \\
- Provide the complete, executable Python code with your modifications, not just the changed parts \\
\\
\#\# Example (Omitting the Chart Image for Brevity) \\
\\
\#\# Example Input \\
\\
**Plotting Code:**
\begin{verbatim}
```python
<example_original_code>
```
\end{verbatim}

**Question:** \\
What is the title of the first subplot on the left? \\
 \\
**Current Answer:** \\
The title of the first subplot is `Sculpture Wave Patterns'. \\
 \\
\#\# Example Output \\
 \\
**Feasibility:** \\
YES \\
\\
**Rationale of Modification:** \\
To change the title of the first subplot, we only need to modify the \texttt{\`{}}ax1.set\_title()\texttt{\`{}} function that sets the title of the first subplot. This change will directly affect the current answer without impacting any other part of the code or plot. Changing the title satisfies the requirement of producing a visually noticeable difference. \\
 \\
**Modified Code:**
\begin{verbatim}
```python
<example_modified_code>
```
\end{verbatim}

**New Answer:** \\
The title of the first subplot is `Dynamic Wave Effects'. \\
 \\
\#\# Input \\
 \\
**Plotting Code:**
\begin{verbatim}
```python
{{ python_code }}
```
\end{verbatim}
**Question:** \\
\{\{ question \}\} \\
 \\
**Current Answer:** \\
\{\{ current\_answer \}\} \\
 \\
\#\# Output Format \\
 \\
**Feasibility:** \\
{[}YES or NO - whether this task can be reasonably accomplished{]} \\
 \\
**Rationale of Modification:** \\
{[}If feasibility is YES: Briefly explain which element(s) you will modify and why this produces a different answer{]} \\
{[}If feasibility is NO: Briefly explain why{]} \\
 \\
**Modified Code:**
\begin{verbatim}
```python
[Your complete modified code here if feasible, otherwise write "None"]
```
\end{verbatim}
**New Answer:** \\
{[}The new correct answer if feasible, otherwise write "None". Do NOT include words like "modified", "updated", "changed", or any reference to the modification process.{]} \\
 \\
\end{tcolorbox}
\caption{The prompt used for descriptive questions.}
\label{fig:descriptive_prompt}
\end{figure*}

\begin{figure*}[t]
\begin{tcolorbox}[
    colback=gray!10,
    colframe=gray!40,
    verbatim,
    fontupper=\footnotesize,
    ]
\scriptsize
**Task:** Given a chart image, its plotting code, a reasoning question, and the current answer with reasoning process, modify the code so that the answer becomes different. You should ONLY modify the element(s) directly responsible for the current answer. \\
\\
\#\# Requirements \\
\\
- First assess whether you think you are capable of reasonably accomplishing this task \\
- Identify the specific data point(s) or element(s) that determine the current answer \\
- Modify ONLY those necessary elements to produce a different answer with a reasoning process \\
- Do NOT change any other data points, labels, colors, or visual elements \\
- Do NOT change the final output/save path in the original code: it must remain `rendered\_images/\{6-digit-number\}.png', e.g., `rendered\_images/000002.png'. \\
- Do NOT modify the \texttt{`{}}set\_random\_seed\texttt{\`{}} function or the random seed value it sets \\
- Ensure the modification is visually noticeable to human eyes (e.g., at least 15-25\% change for numerical values) \\
- Provide the complete, executable Python code with your modifications, not just the changed parts \\
\\
\#\# Example (Omitting the Chart Image for Brevity) \\
\\
\#\#\# Example Input \\
\\
**Plotting Code:**
\begin{verbatim}
```python
<example_original_code>
```
\end{verbatim}
**Question:** \\
By how much does the mean revenue decrease from Q1 to Q2? \\
 \\
**Current Answer:** \\
Reasoning Process: The mean revenue for Q1 is 15.3 and for Q2 it is 11.9. The decrease is calculated as 15.3 - 11.9 = 3.4. \\
Answer: 3.4 \\
 \\
\#\#\# Example Output \\
 \\
**Feasibility:** \\
YES \\
\\
**Rationale of Modification:** \\
To change the answer, I will modify the mean revenue values for Q1 and/or Q2 in \texttt{`{}}revenue\_means\texttt{\`{}}. This adjustment will directly change the mean revenue values displayed in the chart without affecting other elements of the visualization. \\
 \\
**Modified Code:**
\begin{verbatim}
```python
<example_modified_code>
```
\end{verbatim}
**New Answer:** \\
Reasoning Process: The mean revenue for Q1 is 19.1 and for Q2 it is 10.2. The decrease is calculated as 19.1 - 10.2 = 8.9. \\
Answer: 8.9 \\
 \\
\#\# Input \\
 \\
**Plotting Code:**
\begin{verbatim}
```python
{{ python_code }}
```
\end{verbatim}
**Question:** \\
\{\{ question \}\} \\
 \\
**Current Answer:** \\
Reasoning Process: \{\{ current\_reasoning\_process \}\} \\
Answer: \{\{ current\_answer \}\} \\
 \\
\#\# Output Format \\
 \\
**Feasibility:** \\
{[}YES or NO - whether this task can be reasonably accomplished{]} \\
 \\
**Rationale of Modification:** \\
{[}If feasibility is YES: Briefly explain which element(s) you will modify and why this produces a different answer{]} \\
{[}If feasibility is NO: Briefly explain why{]} \\
 \\
**Modified Code:**
\begin{verbatim}
```python
[Your complete modified code here if feasible, otherwise write "None"]
```
\end{verbatim}
**New Answer:** \\
Reasoning Process: {[}If feasible, provide step-by-step reasoning that leads to the new answer, Otherwise write "None". Do NOT include words like "modified", "updated", "changed", or any reference to the modification process.{]} \\
Answer: {[}The new correct answer if feasible, otherwise write "None". Do NOT include words like "modified", "updated", "changed", or any reference to the modification process.{]} \\
 \\
\end{tcolorbox}
\caption{The prompt used for reasoning questions.}
\label{fig:reasoning_prompt}
\end{figure*}

\begin{figure*}[t]
\begin{tcolorbox}[
    colback=gray!10,
    colframe=gray!40,
    verbatim,
    fontupper=\footnotesize,
    ]
\scriptsize

You are an expert at evaluating visualization chart plots. You will be given two python-generated chart images:  \\
- **Original Image**: The chart before code modification \\
- **Modified Image**: The chart after code modification \\
Your task is to assess the similarity between the two chart images. \\
 \\
\#\#\# Scoring Criteria: \\
Evaluate the similarity between the two images based on the following criteria, totaling 100 points: \\
 \\
1. **Chart Types (20 points):** How similar are the chart types (e.g., line charts, bar charts, scatter plots, etc.) between the two images? \\
2. **Layout (20 points):** How similar is the arrangement of subplots (e.g., number of rows and columns, spacing) between the two images? \\
3. **Text Content (20 points):** How similar are the titles, annotations, axis labels, and other text elements (excluding axis tick labels) between the two images? \\
4. **Data (20 points):** How closely do the data trends, patterns, and the number of data groups match between the two images? \\
5. **Style (20 points):** How similar are the colors, line styles, marker types, legends, grids, and other stylistic details between the two images? \\
 \\
\#\#\# Evaluation: \\
Compare the two images head to head and provide a detailed assessment. Use the following format for your response: \\
 \\
 \\
--- \\
 \\
Comments: \\
- Chart Types: \{your comment and subscore\} \\
- Layout: \{your comment and subscore\} \\
- Text Content: \{your comment and subscore\} \\
- Data: \{your comment and subscore\} \\
- Style: \{your comment and subscore\} \\
 \\
Score: \{your final score out of 100\} \\
 \\
--- \\
 \\
Please use the above format to ensure the evaluation is clear and comprehensive. \\
 \\
\end{tcolorbox}
\caption{The prompt used for evaluating image similarity scores.}
\label{fig:similarity_prompt}
\end{figure*}

\section{Synthesis Cost and Data Validation}
\label{sec:cost_and_yield}

To provide a complete picture of ChartCF's data synthesis pipeline, we report the API cost, the multi-stage validation pipeline, and the end-to-end yield from seed samples to final training pairs.

\paragraph{API Cost.}
Table~\ref{tab:synthesis_cost} summarizes the cost of each stage in our synthesis pipeline. The total cost for generating 10K counterfactual pairs is approximately \$427, roughly 1/5 of the \$2,145 reported by ECD~\cite{yang2025effective} for its full dataset. We note that this cost can be further reduced in several ways: 

\begin{itemize}
    \item (1) As shown in Figure~\ref{fig:3}, synthesizing only $\sim$4K pairs (without the filtering step) already matches ECD's performance on CharXiv, reducing the synthesis cost to approximately \$162 ($\sim$1/13 of ECD's cost).
    \item (2) The relatively high cost is partly due to GPT-5's reasoning tokens. Replacing GPT-5 with GPT-4o as the code modifier yields comparable performance at substantially lower cost, as shown in Section \ref{sec:different_code_modifiers}.
    \item (3) Open-source models (e.g., Qwen3-VL-8B-Thinking) can also serve as effective code modifiers, allowing practitioners to run the code modification step locally and avoid API costs entirely.
\end{itemize}

Even accounting for the \$71 cost of generating the 10K ECD seed samples we build upon, our full end-to-end pipeline totals \$499 ($\sim$1/4 of ECD's cost) while requiring significantly less training data (4K preference pairs vs. 300K SFT samples).

\begin{table}[t]
\small
\centering
\resizebox{\columnwidth}{!}{
\begin{tabular}{llr}
\toprule
Stage & Model & Cost (USD) \\
\midrule
Counterfactual code generation (10K samples) & GPT-5 & \$403.93 \\
Similarity scoring (10K pairs) & GPT-5-mini & \$23.38 \\
\midrule
\textbf{ChartCF synthesis total} & & \textbf{\$427.31} \\
\bottomrule
\end{tabular}
}
\caption{\label{tab:synthesis_cost}API cost breakdown of ChartCF's data synthesis pipeline.}
\end{table}

\paragraph{Data Validation.}

To ensure the quality of the synthesized counterfactual data, we adopt a multi-stage validation pipeline:

\begin{itemize}
    \item \textbf{Stage 1: Feasibility check.} Our prompts (Figures~\ref{fig:descriptive_prompt} and~\ref{fig:reasoning_prompt}) require the code modifier to first assess whether a meaningful modification is feasible before generating any code. Infeasible cases are filtered out at this stage. In practice, only 8 out of 10{,}512 seed samples ($\sim$0.08\%) are deemed infeasible.
    \item \textbf{Stage 2: Code parsing and execution.} The modified code is first parsed. Approximately 160 samples ($\sim$1.5\%) fail at this stage. Successfully parsed code is then executed to render the counterfactual chart image, with approximately 832 samples ($\sim$8\%) failing to render.
    \item \textbf{Stage 3: Retry mechanism.} For samples that fail at any of the above stages, we retry the code modifier up to two additional times. After retries, only 84 out of 10{,}512 samples ($\sim$0.8\%) ultimately fail to produce a valid counterfactual pair, yielding 10{,}428 valid pairs (99.2\% success rate).
    \item \textbf{Stage 4: Human verification.} We further randomly sample 100 counterfactual pairs generated by GPT-5 (50 descriptive and 50 reasoning) for manual inspection. For each pair, annotators check: (i) whether the code modification follows the prompt instructions and targets only answer-critical elements, (ii) whether the rendered counterfactual chart is visually plausible, and (iii) whether the new answer $A_{c}$ is factually correct and can be reasonably inferred from the modified chart alone.
    Results show that 98 out of 100 pairs pass inspection (49/50 for descriptive and 49/50 for reasoning questions). While not perfect, this level of quality is sufficient for effective preference learning, as demonstrated by the performance improvements observed across five benchmarks in our experiments.
\end{itemize}

Finally, the similarity-based selection retains roughly 40\% of the valid pairs ($\sim$4K) for training.
We also note that the similarity-based data selection strategy (Section~\ref{subsec:data_selection_strategy}) serves as an additional implicit quality filter, as poorly constructed counterfactual pairs with unintended large visual differences tend to be filtered out.

\section{Benchmark Details}
\label{sec:benchmark_details}

We conduct experiments across five widely-adopted benchmarks, encompassing both real-world and synthetic charts.

\paragraph{Real-world Benchmarks.}
\textbf{CharXiv} \cite{DBLP:conf/nips/WangXH0LZLWLMCA24} contains 2,323 charts sourced from scientific literature, with 4,000 descriptive questions targeting basic chart element recognition and 1,000 reasoning questions requiring high-level reasoning across diverse chart elements.
\textbf{ChartQA} \cite{DBLP:conf/acl/MasryLTJH22} consists of 1,509 charts collected from 4 online sources and 2,500 question-answer pairs.

\paragraph{Synthetic Benchmarks.}
\textbf{ChartBench} \cite{xu2024chartbenchbenchmarkcomplexvisual} provides 2,100 charts with 16,800 ``yes/no'' questions and 2,100 numerical questions.
\textbf{ChartX} \cite{DBLP:journals/tip/XiaYYLZCSYZ25} contains 1,152 test samples across various chart types.
\textbf{ECDBench} \cite{yang2025effective} includes 1,224 chart images, each accompanied by one descriptive question and one reasoning question.

\section{Curriculum Learning for Hard Samples}
\label{sec:curriculum}

To investigate whether curriculum learning could better leverage difficult samples, we design a two-stage training strategy. In the first stage, models are trained on low-similarity pairs (easier samples). In the second stage, we continue training with a reduced learning rate (1e-5) on either high-similarity pairs or data with distractors (see Section \ref{subsec:different_types_of_hard_samples} for details about these hard sample types).
Our motivation stems from the finding that directly training on overly difficult samples degrades performance. We hypothesize that starting with easier samples before progressively introducing harder ones might enable better handling of challenging cases.

Results in Table~\ref{tab:curriculum_learning} show that curriculum learning fails to outperform the simpler strategy of using only low-similarity pairs. 
This suggests that in this setting, careful data selection is more effective than curriculum-based schedules.
However, we believe there may exist more effective strategies to leverage such hard samples, a direction we plan to explore in future work.

\begin{table}[t]
\small
\centering
\resizebox{\columnwidth}{!}{
\begin{tabular}{lccc}
\toprule
\multirow{2}{*}{Method} & \multicolumn{3}{c}{CharXiv}\\
\cmidrule{2-4}
 & Des. & Rea. & Avg. \\ 
\midrule
Qwen2.5-VL & 66.40 & 41.20 & 61.36 \\
\hdashline
\multicolumn{4}{l}{\textcolor{gray}{\textit{Synthetic Data w/o Distractors}}} \\
ChartCF w/o Data Selection (10K) & 73.08 & 43.20 & 67.10 \\
\ \ \ \ w. Lowest Half Pairs (5K) & \textbf{75.65} & \textbf{43.70} & \textbf{69.26} \\
\ \ \ \ w. Highest Half Pairs (5K) & 66.42 & 43.30 & 61.80 \\
\hdashline
\multicolumn{4}{l}{\textcolor{gray}{\textit{Curriculum Learning}}} \\
\ \ \ \ w. Stage-1: L (5K), Stage-2: H (5K) & 75.60 & 42.80 & 69.04 \\
\ \ \ \ w. Stage-1: L (5K), Stage-2: Dis (5K) & 75.20 & 42.60 & 68.68 \\
\bottomrule
\end{tabular}
}
\caption{\label{tab:curriculum_learning}Curriculum learning experiments on CharXiv. ``L'', ``H'', and ``Dis'' denote ``Lowest Half Pairs'', ``Highest Half Pairs'', and data with distractors, respectively.}
\end{table}

\section{Case Study of Counterfactual Data Synthesis}
\label{sec:case_study_of_counterfactual_data_synthesis}

We present four representative examples of counterfactual chart pairs generated by our synthesis pipeline, covering both descriptive and reasoning questions with varying modification types and similarity scores.

\begin{figure*}
\centering
\includegraphics[width=0.95\textwidth]{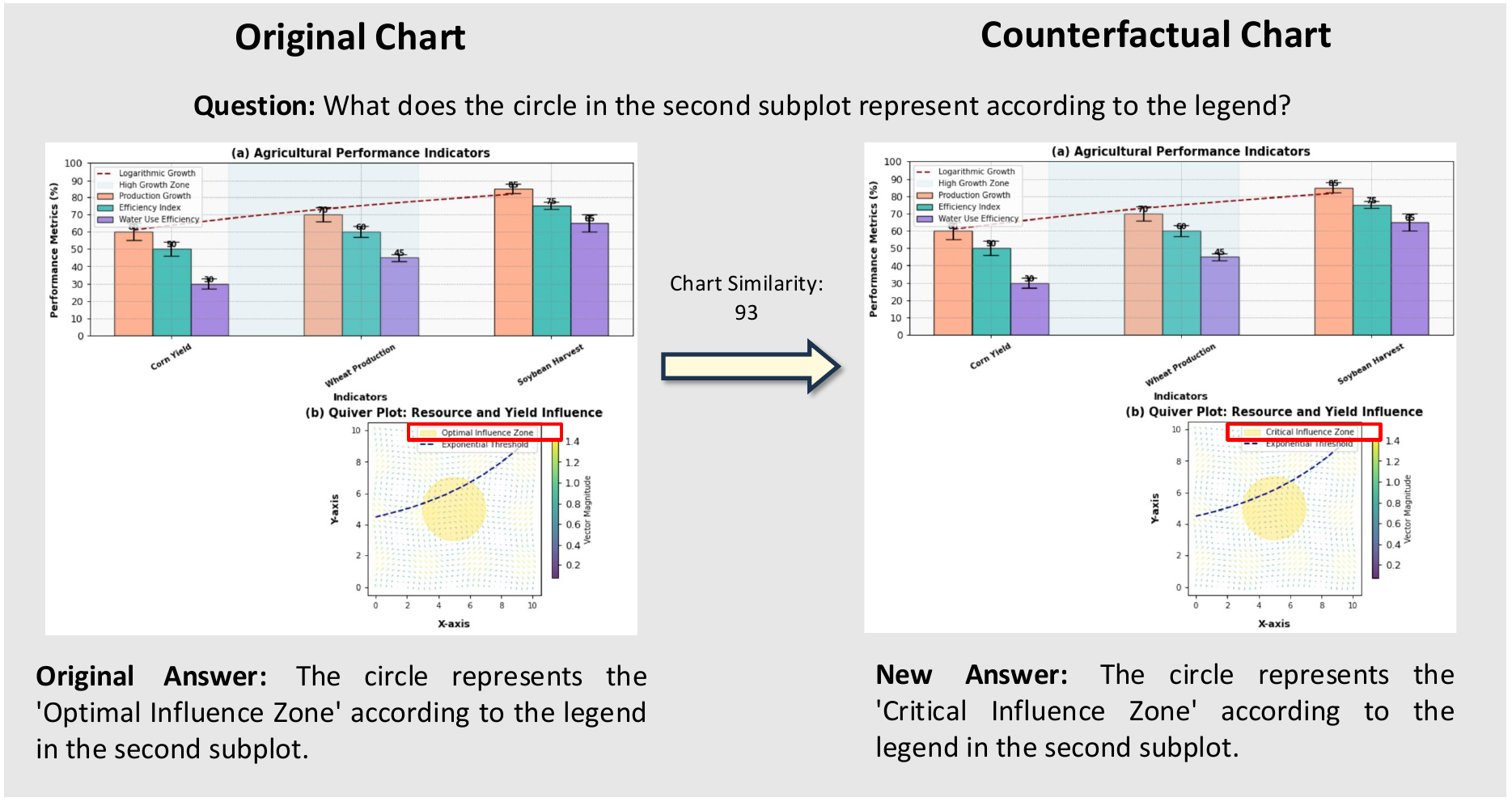}
\caption{\label{case_data_synthesis:1}Counterfactual pair for a descriptive question about legend interpretation. The legend label ``Optimal Influence Zone'' is modified to ``Critical Influence Zone'' (similarity: 93).}
\end{figure*}

\begin{figure*}
\centering
\includegraphics[width=0.95\textwidth]{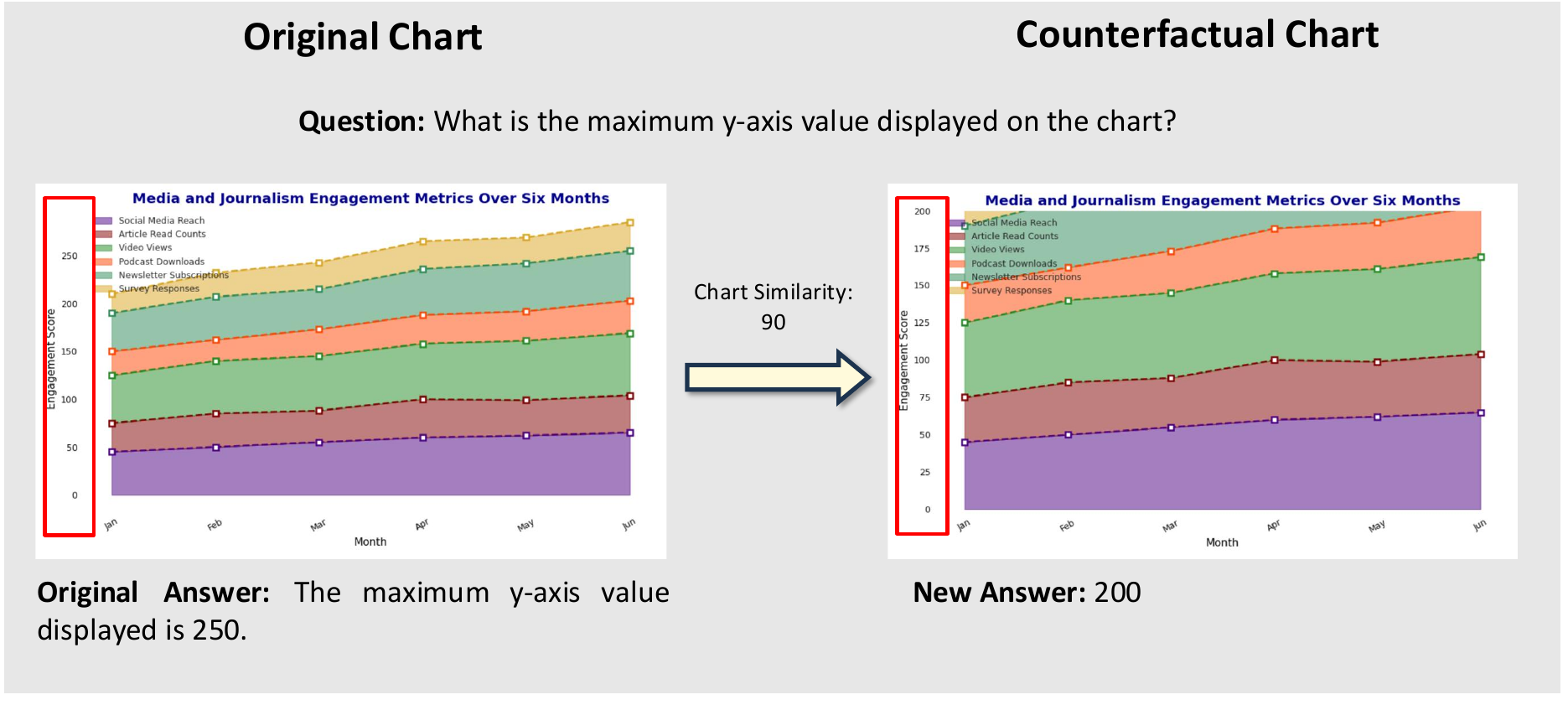}
\caption{\label{case_data_synthesis:2}Counterfactual pair for a descriptive question about y-axis scale. The maximum y-axis value is changed from 250 to 200 (similarity: 90).}
\end{figure*}

\begin{figure*}
\centering
\includegraphics[width=0.95\textwidth]{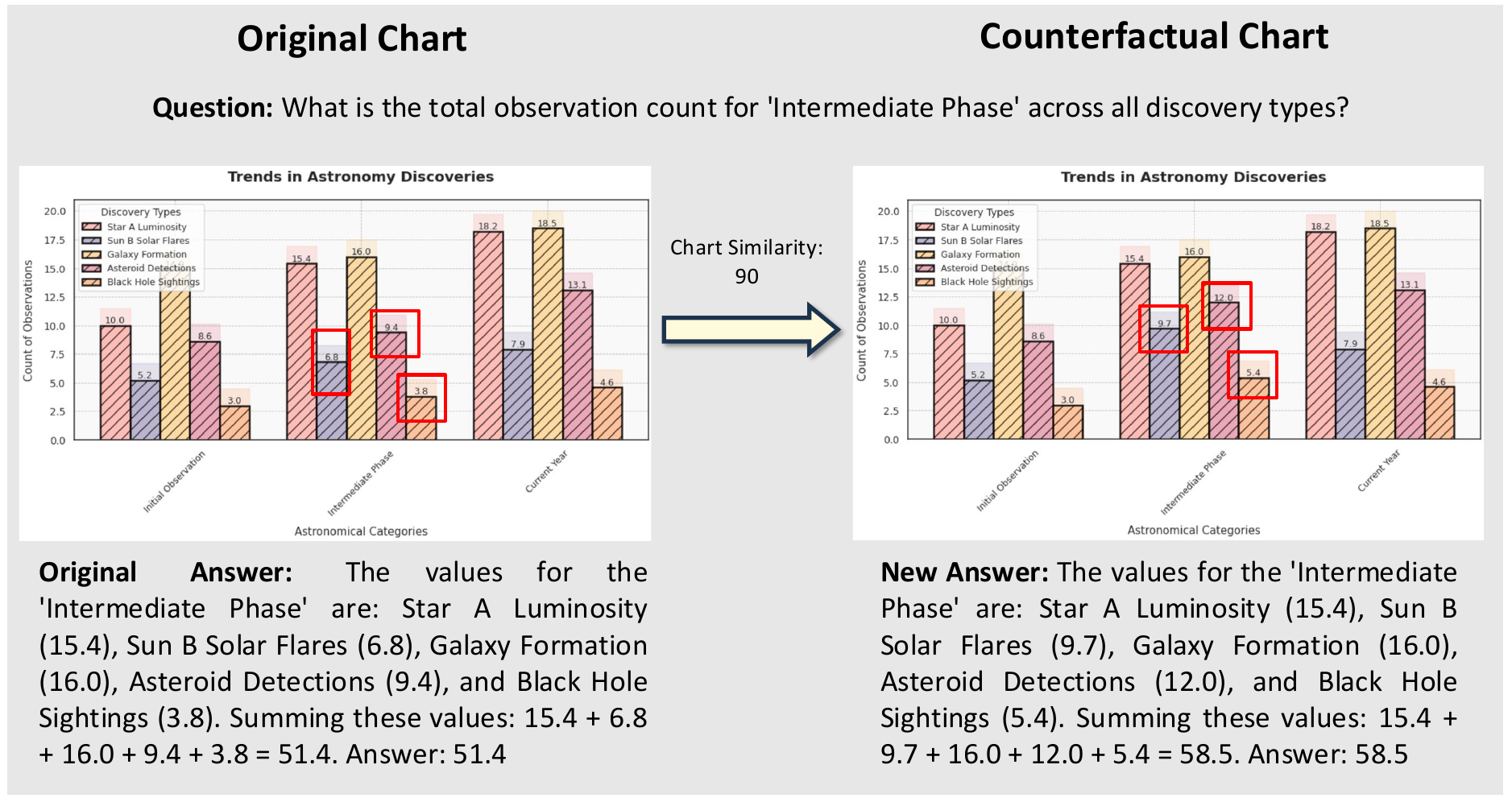}
\caption{\label{case_data_synthesis:3}Counterfactual pair for a reasoning question involving numerical computation. Multiple data values are modified, resulting in a different sum (similarity: 90).}
\end{figure*}

\begin{figure*}
\centering
\includegraphics[width=0.95\textwidth]{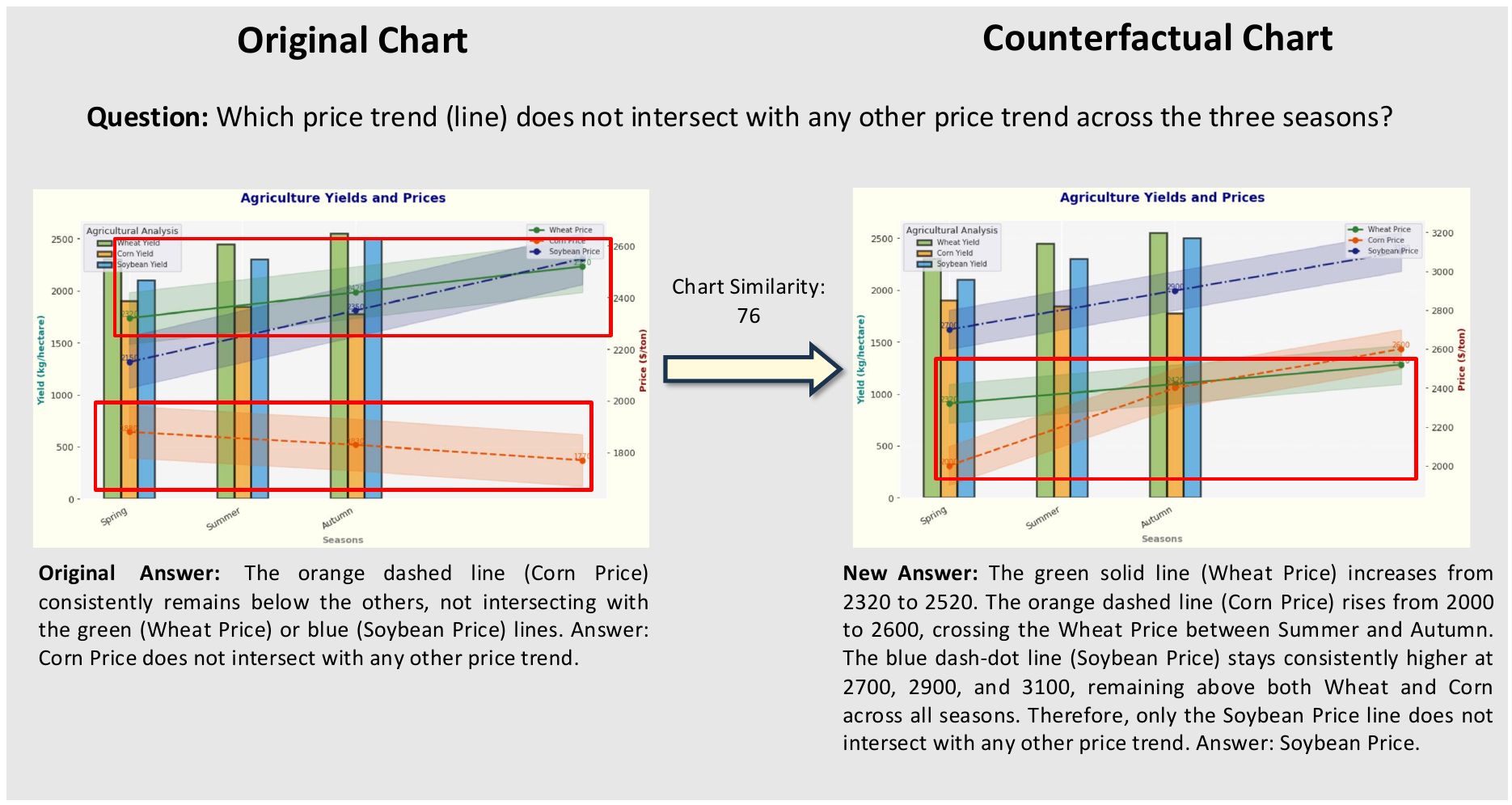}
\caption{\label{case_data_synthesis:4}Counterfactual pair for a reasoning question about line intersections. The Corn Price trend is modified from declining to increasing, changing spatial relationships (similarity: 76).}
\end{figure*}

\begin{enumerate}
    \item \textbf{Case 1} (Figure~\ref{case_data_synthesis:1}, similarity: 93) illustrates a descriptive question requiring legend interpretation. Our pipeline modifies the textual label in the legend from ``Optimal Influence Zone'' to ``Critical Influence Zone'' while keeping all visual elements unchanged, creating a relatively high-similarity counterfactual pair.
    \item \textbf{Case 2} (Figure~\ref{case_data_synthesis:2}, similarity: 90) illustrates a descriptive question about axis scale recognition. The modification changes the maximum y-axis value from 250 to 200.
    \item \textbf{Case 3} (Figure~\ref{case_data_synthesis:3}, similarity: 90) shows a reasoning question involving value extraction and numerical computation. Multiple data values are modified (e.g., Sun B Solar Flares from 6.8 to 9.7), resulting in a different final sum while maintaining similar chart appearance.
    \item  \textbf{Case 4} (Figure~\ref{case_data_synthesis:4}, similarity: 76) presents a reasoning question about spatial relationships between chart objects. The modification alters line positions, requiring identification of both visual elements and their relative positions.
\end{enumerate}

These examples demonstrate our pipeline's ability to create diverse counterfactual pairs through different modification strategies: altering numerical values (Cases 2, 3), modifying textual information (Case 1), and changing visual element positions (Case 4).

\section{Case Study of Model Predictions}
\label{sec:case_study_of_model_predictions}

We compare predictions between ChartCF and the ECD baseline (both based on Qwen2.5-VL) on five challenging examples from CharXiv, demonstrating how our method enables more fine-grained understanding of chart details.

\begin{figure*}
\centering
\includegraphics[width=0.95\textwidth]{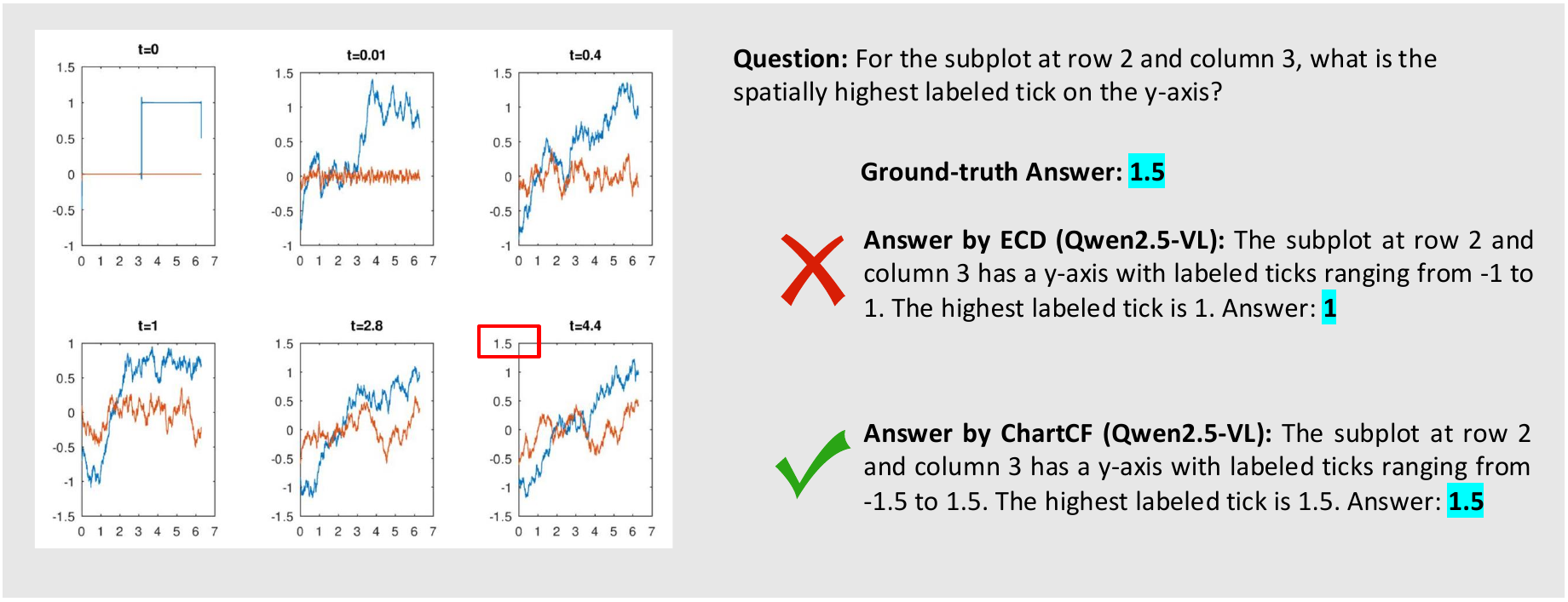}
\caption{\label{case_model_prediction:1}Comparison on a descriptive question about y-axis tick values. ECD misidentifies the subplot location and axis range, while ChartCF correctly identifies the target subplot and reads the highest tick value.} 
\end{figure*}

\begin{figure*}
\centering
\includegraphics[width=0.95\textwidth]{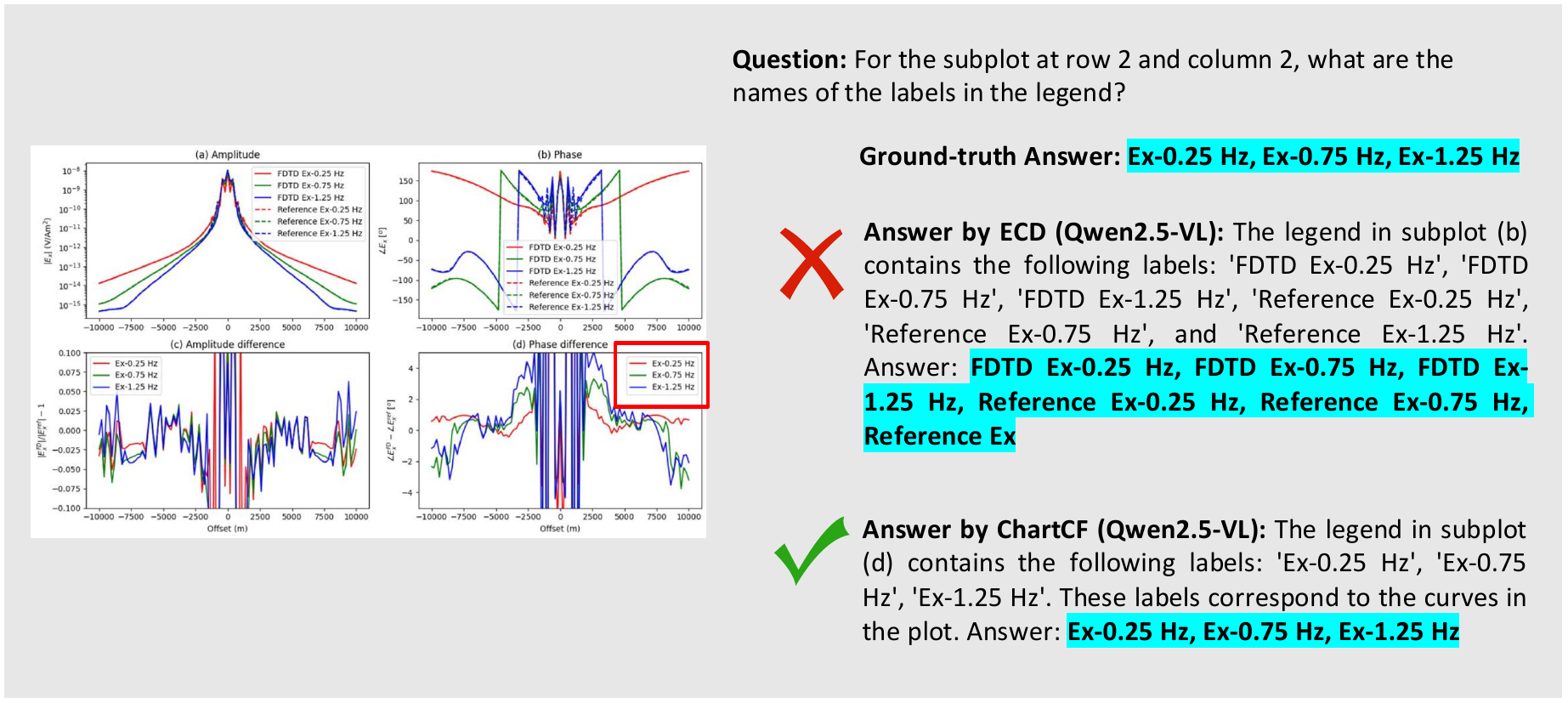}
\caption{\label{case_model_prediction:2}Comparison on a descriptive question about legend labels. ECD extracts labels from the wrong subplot, while ChartCF accurately locates the correct subplot and extracts the relevant labels.} 
\end{figure*}

\begin{figure*}
\centering
\includegraphics[width=0.95\textwidth]{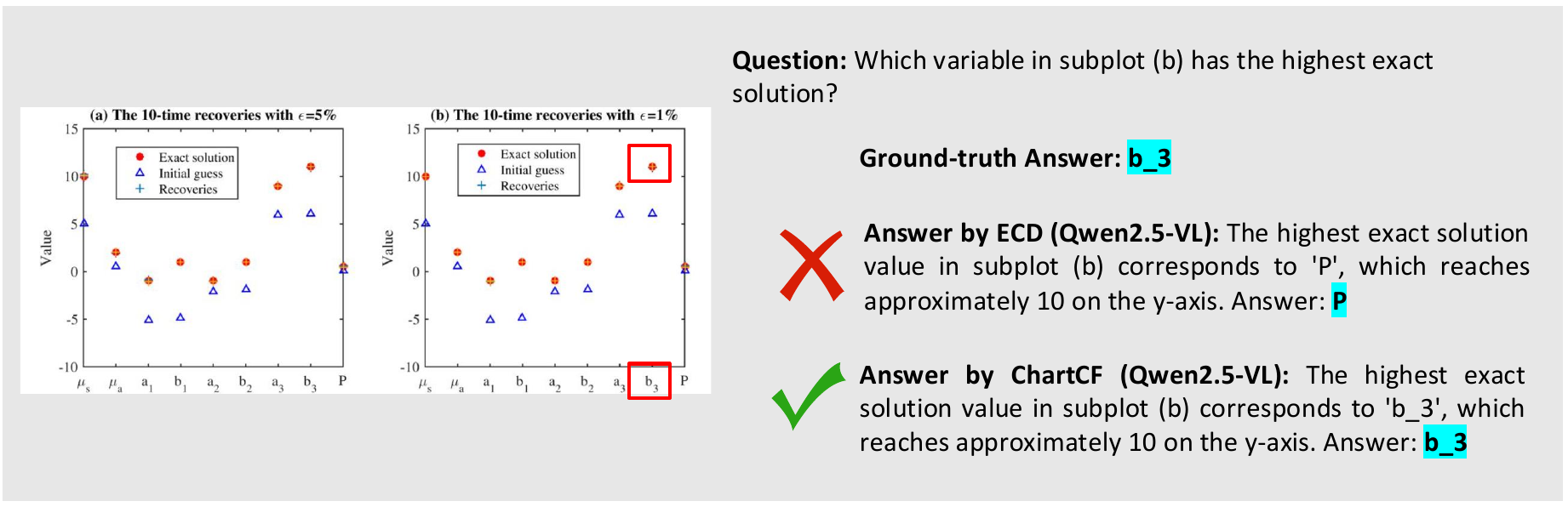}
\caption{\label{case_model_prediction:3}Comparison on a reasoning question about identifying the highest solution value. ECD confuses similar visual markers, while ChartCF correctly identifies ``b\_3'' through precise spatial reasoning.} 
\end{figure*}

\begin{figure*}
\centering
\includegraphics[width=0.95\textwidth]{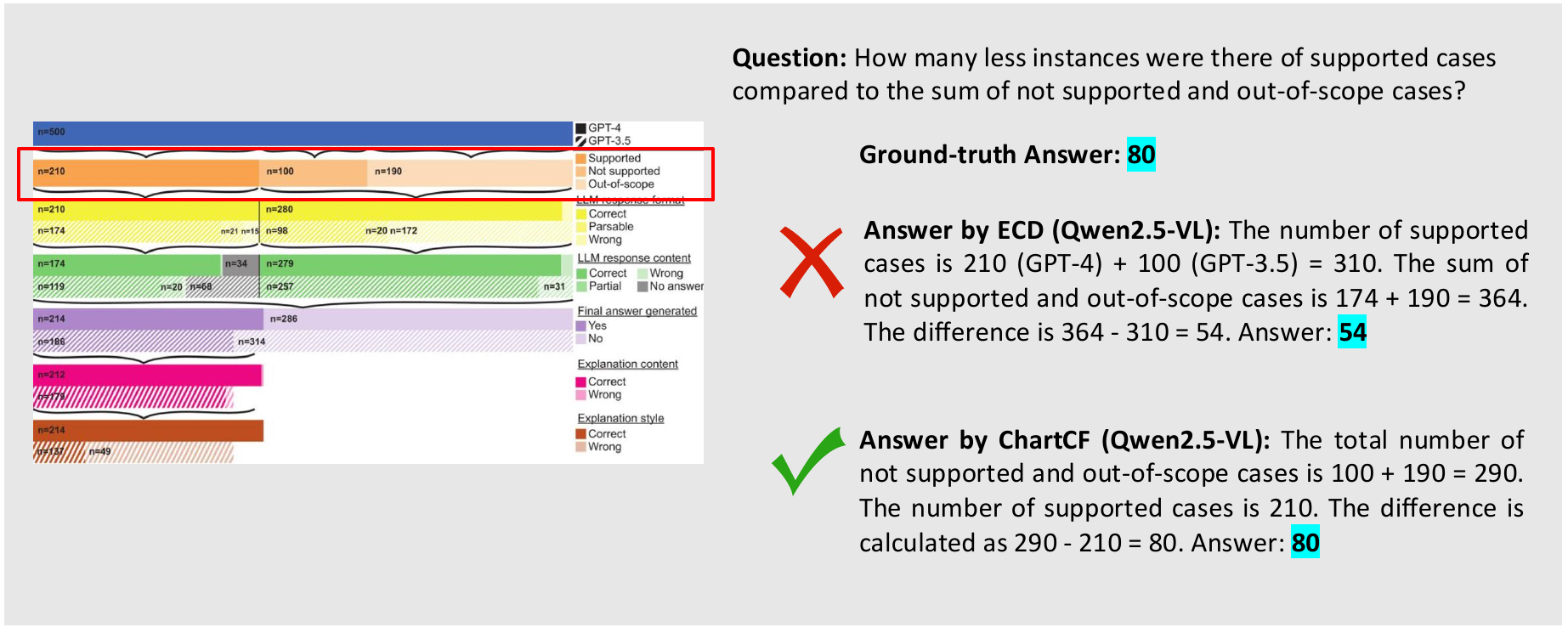}
\caption{\label{case_model_prediction:4}Comparison on a reasoning question involving numerical computation from legend categories. ECD makes calculation errors due to misreading counts, while ChartCF performs accurate value extraction and computation.} 
\end{figure*}

\begin{figure*}
\centering
\includegraphics[width=0.95\textwidth]{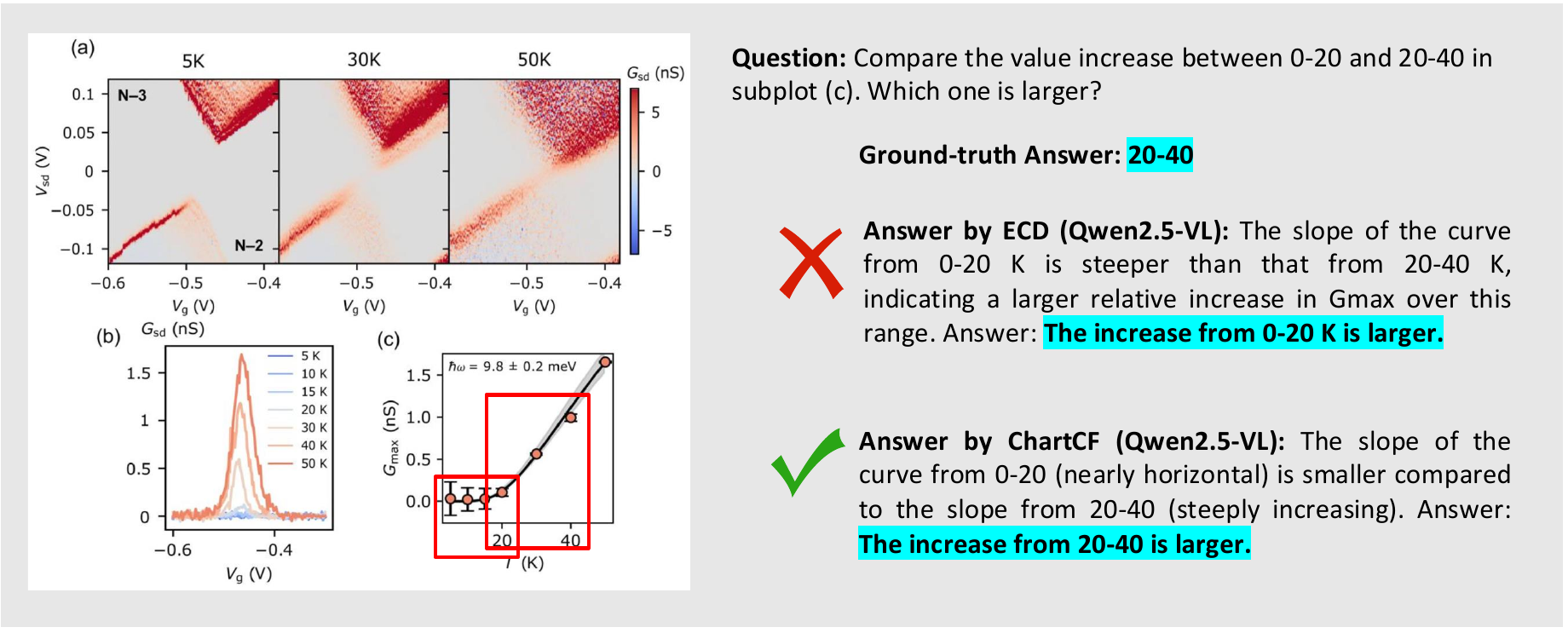}
\caption{\label{case_model_prediction:5}Comparison on a reasoning question about comparing curve slopes across regions. ECD incorrectly interprets the slope comparison direction, while ChartCF accurately analyzes curve steepness to determine the larger increase.} 
\end{figure*}

\begin{enumerate}
    \item \textbf{Case 1} (Figure~\ref{case_model_prediction:1}) involves a descriptive question requiring precise identification of a specific subplot and accurate reading of small-scale y-axis tick values. ECD fails to correctly identify the subplot location and misreads the axis range, while ChartCF correctly locates the target subplot and identifies the highest tick value as 1.5.
    
    \item \textbf{Case 2} (Figure~\ref{case_model_prediction:2}) presents a descriptive question about legend label extraction from a specific subplot. ECD incorrectly identifies the subplot and extracts labels from the wrong legend. ChartCF accurately locates the correct subplot and extracts only the relevant labels.
    
    \item \textbf{Case 3} (Figure~\ref{case_model_prediction:3}) demonstrates a reasoning question involving spatial position understanding. ECD misidentifies the target variable by confusing similar visual markers, while ChartCF correctly identifies ``b\_3'' through precise spatial reasoning.
    
    \item \textbf{Case 4} (Figure~\ref{case_model_prediction:4}) shows a reasoning question requiring numerical computation from legend values. ECD makes calculation errors by misreading category counts, while ChartCF performs accurate value extraction and computation to reach the correct answer.
    
    \item \textbf{Case 5} (Figure~\ref{case_model_prediction:5}) illustrates a reasoning question about comparing visual patterns across subplot regions. ECD incorrectly interprets the slope comparison, while ChartCF accurately analyzes the curve steepness in different regions to determine the correct answer.
\end{enumerate}

These cases highlight that ChartCF's counterfactual training enables models to develop more precise visual discrimination capabilities, particularly in tasks requiring fine-grained element localization, accurate value reading, spatial relationship understanding, and multi-step reasoning over chart elements.

\end{document}